\documentclass{article}

\usepackage[preprint]{arxiv} 

\usepackage{geometry}
\usepackage{graphicx}
\usepackage{subcaption}

\usepackage{caption}
\usepackage{enumitem}

\usepackage{amsfonts}
\usepackage{amssymb}
\usepackage{amsmath}
\usepackage{amsthm}
\usepackage{bm}
\usepackage[thicklines]{cancel}
\usepackage[ruled]{algorithm2e}
\usepackage{cleveref}

\title{A Practical Introduction to Deep Reinforcement Learning}

\author{
    Yinghan Sun$^{1}$, Hongxi Wang$^{1}$, Hua Chen$^{2}$, Wei Zhang$^{1}$\\
    $^{1}$Southern University of Science and Technology\\
    $^{2}$Zhejiang University-University of Illinois Urbana-Champaign Institute\\
    \texttt{sunyh2021@sustech.edu.cn} \\
}

\begin{document}
\maketitle


\begin{abstract}
Deep reinforcement learning (DRL) has emerged as a powerful framework for solving sequential decision-making problems, achieving remarkable success in a wide range of applications, including game AI, autonomous driving, biomedicine, and large language models. However, the diversity of algorithms and the complexity of theoretical foundations often pose significant challenges for beginners seeking to enter the field. This tutorial aims to provide a concise, intuitive, and practical introduction to DRL, with a particular focus on the Proximal Policy Optimization (PPO) algorithm, which is one of the most widely used and effective DRL methods. To facilitate learning, we organize all algorithms under the Generalized Policy Iteration (GPI) framework, offering readers a unified and systematic perspective. Instead of lengthy theoretical proofs, we emphasize intuitive explanations, illustrative examples, and practical engineering techniques. This work serves as an efficient and accessible guide, helping readers rapidly progress from basic concepts to the implementation of advanced DRL algorithms.
\end{abstract}

\vspace{10mm}
\begin{figure}[htbp]
    \centering
    \includegraphics[width=0.85\textwidth, trim = 0 0 35 0, clip]{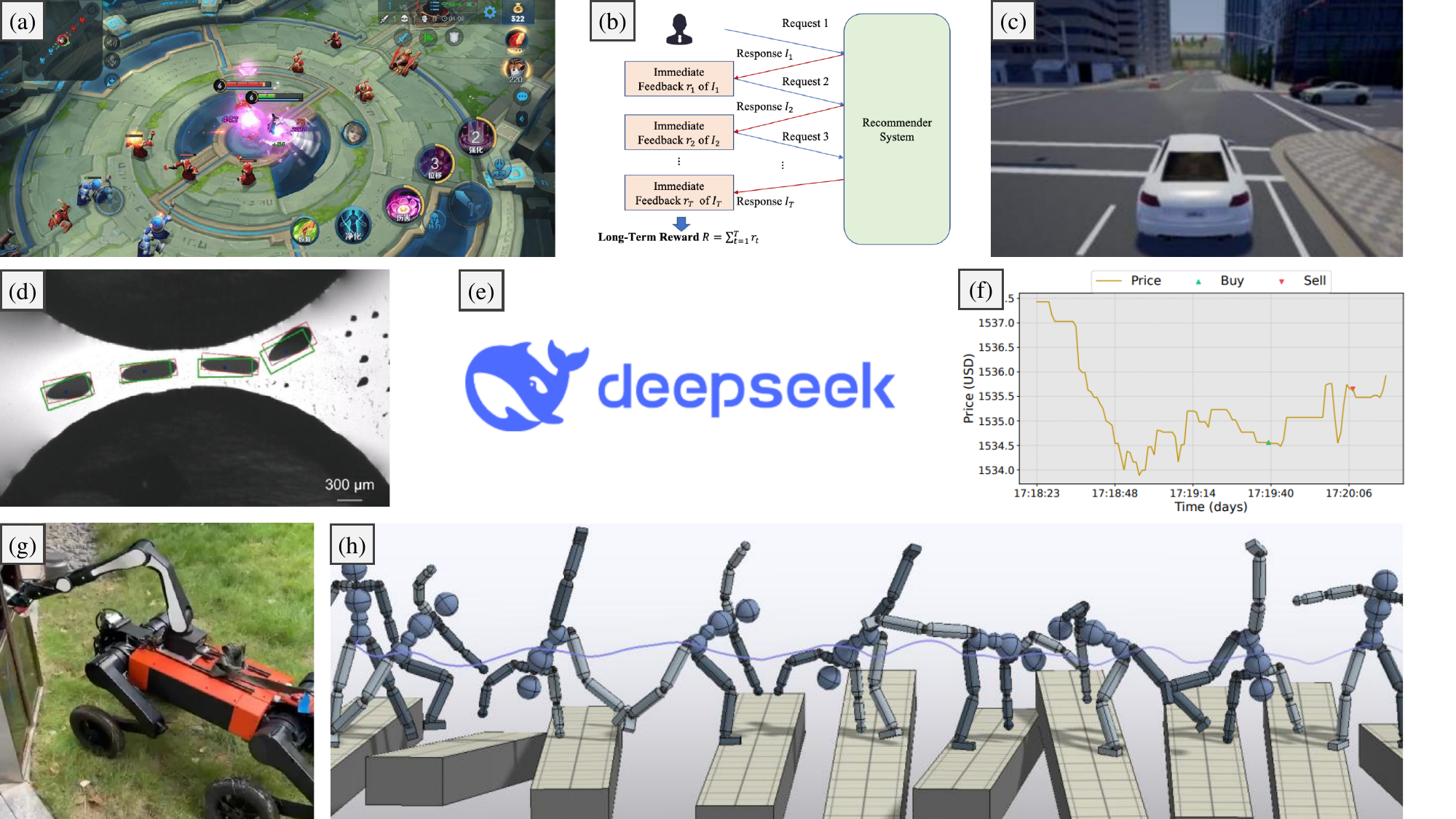}

    \caption{\textbf{Reinforcement learning has been extensively applied to a wide range of domains.} (a) Game AI~\cite{rl_applications_game_ai}. (b) Recommendation systems~\cite{rl_applications_recommendation_sys}. (c) Autonomous driving~\cite{rl_applications_autonomous_driving}. (d) Biomedicine~\cite{rl_applications_biomedicine}. (e) Large language models (LLMs)~\cite{rl_applications_llm}. (f) Quantitative trading~\cite{rl_applications_quantitative_trading}. (g) Robotics~\cite{rl_applications_robotics}. (h) Computer animation~\cite{rl_applications_computer_animation}.}

    \label{fig:rl_applications}
\end{figure}


\newpage

\tableofcontents

\newpage


\section{Introduction}

Reinforcement learning (RL) has emerged as a fundamental paradigm for sequential decision-making, enabling agents to learn optimal behaviors through interactions with dynamic environments. The integration of deep neural networks with reinforcement learning has given rise to deep reinforcement learning (DRL), which further revolutionizes the field by facilitating end-to-end learning of both perception and control policies in high-dimensional state spaces. As illustrated in Figure~\ref{fig:rl_applications}, DRL has progressed from theoretical foundations to transformative applications across diverse domains, including game AI~\cite{rl_applications_game_ai}, recommendation systems~\cite{rl_applications_recommendation_sys}, autonomous driving~\cite{rl_applications_autonomous_driving}, biomedicine~\cite{rl_applications_biomedicine}, large language models (LLMs)~\cite{rl_applications_llm}, quantitative trading~\cite{rl_applications_quantitative_trading}, robotics~\cite{rl_applications_robotics}, and computer animation~\cite{rl_applications_computer_animation}. This growing impact has sparked widespread interest in learning DRL.

Despite its promise, many beginners often feel overwhelmed by the large variety of reinforcement learning algorithms, discouraged by intricate theoretical proofs, or left with only a superficial grasp of key methods without a deeper understanding of the underlying principles. This tutorial aims to bridge that gap by providing a fast track from foundational concepts to advanced algorithms in reinforcement learning within a concise, focused scope. In particular, we concentrate on the Proximal Policy Optimization (PPO) algorithm, one of the most practical and widely adopted DRL algorithms. To this end, the tutorial is designed according to the following key principles:
\begin{itemize}
    \item \textbf{Efficiency-Driven:} We intentionally exclude algorithms irrelevant to the primary objective, ensuring readers can efficiently acquire essential DRL concepts and ideas. 
    \item \textbf{Unified Framework:} All algorithms are presented within the Generalized Policy Iteration (GPI) framework, helping readers quickly build a systematic understanding of DRL and recognize the connections between different approaches. 
    \item \textbf{Intuition First:} Instead of lengthy and intricate theoretical proofs, the tutorial prioritizes examples and intuitive explanations, allowing readers to grasp the core ideas and motivations behind algorithm design. 
    \item \textbf{Practical Orientation:} We comprehensively present the fundamental concepts, algorithmic details, and necessary mathematical derivations. Additionally, practical engineering techniques for implementing these algorithms are included, enabling readers to apply reinforcement learning to their own projects. 
\end{itemize}

The remainder of this tutorial is organized as follows. Section~\ref{sec:basic_concepts} introduces the fundamental concepts of reinforcement learning, including the agent-environment interface and value functions. Section~\ref{sec:big_picture} offers a high-level overview of reinforcement learning algorithms under the GPI framework, providing a unified perspective for understanding DRL. Sections~\ref{sec:value_estimation} and~\ref{sec:policy_gradients} delve into two major categories of DRL algorithms: value-based and policy-based methods. Specifically, Section~\ref{sec:value_estimation} explores various value estimation techniques, which are essential components of PPO. Section~\ref{sec:policy_gradients} introduces policy gradient methods and Generalized Advantage Estimation (GAE), two additional core elements of PPO. Finally, we present the complete procedure of the PPO algorithm.
\section{Basic Concepts in Reinforcement Learning}
\label{sec:basic_concepts}

The theory of \textbf{\textit{reinforcement learning (RL)}} provides a framework for understanding how humans and other animals, viewed as agents, can optimize their behavior in natural environments. It serves as a fundamental tool for tackling decision-making problems, where an agent must determine its actions based on observations of the environment to achieve specific goals. In practice, these decision-making problems are often formulated using \textbf{\textit{Markov Decision Processes (MDPs)}}~\cite{rl_an_intro_sutton_2018}, and RL algorithms are subsequently applied to identify the optimal policy for achieving those goals. This section will introduce the key concepts in reinforcement learning.
\begin{figure}[htbp]
\centering
\includegraphics[width=\textwidth, page=2, trim = 80 250 50 40, clip]{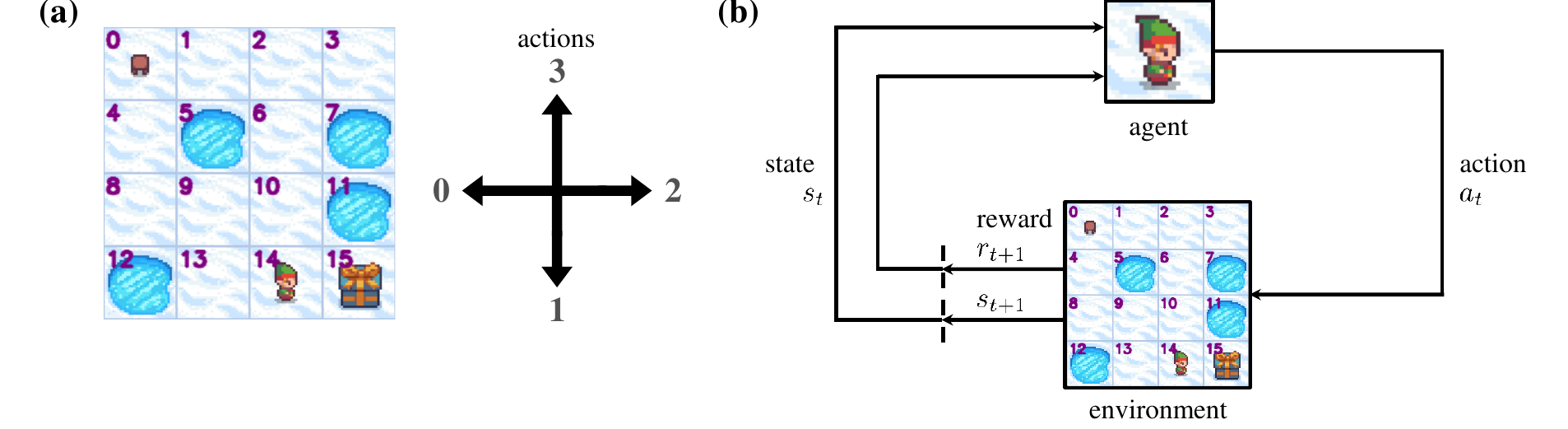}
\caption{\textbf{The agent-environment interface.} (a) A detailed representation of the frozen lake environment. Numbers in purple denote the states. The agent always starts at $s_0 = 0$. The episode ends when the agent reaches a hole ($s_t \in \{5, 7, 11, 12\}$) or the goal state ($s_t = 15$). In each non-terminal state, the agent can choose from four actions: move left, move down, move right, or move up, encoded as 0, 1, 2, and 3, respectively. (b) An illustration of the agent-environment interaction. At each time step $t$, the agent observes the state $\bm{s}_t$, selects an action $\bm{a}_t$ based on this observation, transitions to the next state $\bm{s}_{t+1}$, and receives a reward signal $r_{t+1}$.}
\label{fig:agent_env_interface}
\vspace{-6mm}
\end{figure}

\subsection{Interaction Between Agent and Environment}
We begin our introduction with a toy example in which an agent needs to cross a frozen lake from its starting point to the goal while avoiding falling into any holes~\cite{gymnasium_arxiv2024}. As illustrated in Figure~\ref{fig:agent_env_interface}, the agent begins in an initial state $\bm{s}_0$. At each time step $t$, the agent observes the current \emph{\textbf{state}} $\bm{s}_t$ of the environment and selects an \emph{\textbf{action}} $\bm{a}_t$ based on this observation. One time step later, due to its action, the agent transitions to a new state $\bm{s}_{t+1}$. Additionally, the environment provides the agent with a feedback signal $r_{t+1}$, known as the \emph{\textbf{reward}}, which indicates the quality of the action taken. 
The agent interacts with the environment in this way until it reaches a final state $\bm{s}_T$ (for example, the agent may reach the goal) at time step $T$. This entire process, from the initial state to the final state, is defined as an \emph{\textbf{episode}}. An episode encompasses all the actions, states, and rewards experienced by the agent during its journey.
Throughout the episode, we can record a sequence of state-action-reward pairs, referred to as a \emph{\textbf{trajectory}}. A trajectory captures the complete history of the agent's interactions with the environment and is often denoted as $\tau$:
\begin{equation}
    \tau = \{\bm{s}_0, \bm{a}_0, r_1, \bm{s}_1, \bm{a}_1, r_2, \cdots, \bm{s}_{T-1}, \bm{a}_{T-1}, r_T, \bm{s}_T\}
\end{equation}
In reinforcement learning, we aim to find the optimal policy $\pi^{*}$ to maximize the accumulated reward collected along the trajectory. The following provides a detailed explanation of this process using the frozen lake environment as an example.

\subsubsection*{2.1.1~~ State}
We denote the \emph{\textbf{state space}} as $\mathcal{S}$, representing the set of all possible states an agent can occupy. The \textbf{\textit{initial state}}, denoted by $\bm{s}_0$, is the state where an episode begins. This initial state is drawn from a probability distribution $p(\bm{s}_0)$, which specifies the likelihood of starting the episode in each possible state. In some applications, the initial state is deterministic, meaning $\bm{s}_0$ is predefined as a single specific state. We also define \textbf{\textit{terminal states}} as states that end an episode, either due to the agent achieving a goal or encountering a failure. Once the agent reaches a terminal state, no further actions can be taken.  

For example, in the frozen lake environment, a state corresponds to the agent's location, represented by an integer $s_t = 4i + j$, where $i, j \in \{0, 1, 2, 3\}$ are the row and column indices, respectively. The state space can thus be expressed as $\mathcal{S} = \{s \in \mathbb{Z} \mid 0 \leq s \leq 15\}$. In this environment, the initial state is deterministically set to $s_0 = 0$, and the target state is $s = 15$. The terminal states are defined as $\{5, 7, 11, 12, 15\}$, while all other states are non-terminal.  

It is important to carefully define states in practical applications, as their design directly impacts the agent's ability to make informed decisions. In Markov Decision Processes (MDPs), the state $s_t$ must encapsulate sufficient information to enable the agent to select appropriate actions based solely on its current observation. For instance, if $s_t$ only includes the agent's row position ($s_t = i, i \in \{0, 1, 2, 3\}$) but omits its column position, the agent will lack the necessary information to decide lateral movements (left or right). Instead, such a state representation would restrict decisions to vertical movements (up or down), leading to suboptimal behavior.

\subsubsection*{2.1.2~~ Action}
We denote the \emph{\textbf{action space}} by $\mathcal{A}$, which represents the set of all possible actions. In the frozen lake environment, the agent has four available actions: $\mathcal{A} ={ }${\ttfamily \{moving left, moving down, moving right, moving up\}}. To facilitate mathematical representation and coding, we encode these actions using integers. Specifically, we can use $0, 1, 2, 3$ to represent moving left, down, right, and up, respectively.

\subsubsection*{2.1.3~~ Policy}
Mathematically, a \textbf{\textit{policy}} is defined as a conditional probability distribution that specifies the likelihood of selecting each action given a particular state, commonly denoted as $\pi(\bm{a} | \bm{s})$. Policies can be categorized as either deterministic or stochastic. A \textbf{\textit{deterministic policy}} selects a specific action with probability 1, assigning a probability of 0 to all other actions. In contrast, a \textbf{\textit{stochastic policy}} defines a probability distribution over the available actions, introducing randomness into the action selection process. For example, in the frozen lake environment, a deterministic policy might dictate that the agent always chooses the {\ttfamily move right} action, regardless of its current state. Conversely, a stochastic policy could take the form of a random policy, where the agent assigns equal probabilities to all available actions in every state.

\subsubsection*{2.1.4~~ Reward}
During interaction, when the agent selects an action $\bm{a}_t$ in a given state $\bm{s}_t$, it receives a \textbf{\textit{reward}} signal $r_{t+1}$, which evaluates the quality of the action $\bm{a}_t$. A higher reward incentivizes the agent to repeat the action, while a lower reward discourages it. Unlike the natural interactions of humans and animals with their environments, solving decision-making problems in reinforcement learning requires manually designing effective reward functions. The reward function maps states and actions to corresponding reward values. In reinforcement learning, the objective is to find the optimal policy by maximizing the expected cumulative reward. Consequently, the design of the reward function plays a critical role in shaping the optimization objective. A poorly designed reward function may lead to a policy that optimizes unintended outcomes or fails to achieve the desired goals.

In the frozen lake environment, the predefined reward function is structured as follows: when the agent takes an action, it receives a reward of $+1$ upon reaching the goal state; otherwise, the reward is 0.

\subsubsection*{2.1.5~~ Return}
The \emph{\textbf{return}} is typically defined as the weighted sum of all rewards an agent receives from a given time step $t$ onward:
\begin{equation}
    G_t = r_{t+1} + \gamma r_{t+2} + \gamma^2 r_{t+3} + \cdots + \gamma^{T-t-1} r_T = \sum_{k=0}^{T-t-1} \gamma^k r_{t+k+1},
\end{equation}
where $\gamma \in [0, 1]$ is known as the \emph{\textbf{discount rate}}. This return consists of the \textbf{\textit{immediate reward}} $r_{t+1}$ and \textbf{\textit{future rewards}} $\gamma r_{t+2} + \cdots + \gamma^{T-t-1} r_T$. When $\gamma$ is close to 0, the agent places more emphasis on immediate rewards. Conversely, as $\gamma$ approaches 1, the agent gives greater importance to future rewards.  Additionally, we use $G(\tau)$to denote the accumulated discounted rewards collected along trajectory $\tau$.

\subsubsection*{2.1.6~~ Transition}
When the agent takes an action $\bm{a}_t$, it transitions from the current state $\bm{s}_t$ to a subsequent state $\bm{s}_{t+1}$ and receives a reward signal $r_{t+1}$. This process is defined as a \textbf{\textit{transition}}. The probability of such a transition is denoted by $p(\bm{s}_{t+1}, r_{t+1} | \bm{s}_t, \bm{a}_t)$, which describes the likelihood of the agent moving to state $\bm{s}_{t+1}$ and receiving the reward $r_{t+1}$ when taking action $\bm{a}_t$ in state $\bm{s}_t$. This probability is also referred to as the \textbf{\textit{dynamics}} of the MDP, or the \textbf{\textit{model}}. 

To illustrate, consider the frozen lake environment. When the lake is slippery, the agent may not reach the intended next state even after executing a specific action $a_t$. For example, as depicted in the green block of Figure~\ref{fig:relationship_state_action_values}, if the agent is in state $s_t = 14$ and chooses to move right ($a_t = 2$), it may fail to reach the goal state $s_{t+1} = 15$ due to slipping, and instead end up in another state, such as $s_{t+1} = 13$. This behavior is reflected in the probabilities $p(s_{t+1} = 15, r_{t+1} = 1 | s_t = 14, a_t = 2) = 0.8$ and $p(s_{t+1} = 13, r_{t+1} = 0 | s_t = 14, a_t = 2) = 0.1$. On the other hand, if the lake is not slippery, transitions become deterministic, meaning that $p(s_{t+1}, r_{t+1} | s_t, a_t) = 1$ holds for all actions and states.

\subsubsection*{2.1.7~~ Termination of an episode}
Generally, an episode ends when the agent reaches a terminal state. However, in practical applications, waiting for the agent to reach a terminal state may take an excessive amount of time. In some cases, an episode may not even have a terminal state. Since researchers have limited time to conduct experiments, a common approach is to impose a maximum length on episodes. Once this maximum length is reached, the episode ends, even if it could theoretically continue. To distinguish between different reasons for ending an episode, researchers have introduced the concepts of \textbf{\textit{termination}} and \textbf{\textit{truncation}}~\cite{gymnasium_arxiv2024}. Termination refers to a state-based ending, which signifies either the success or failure of the task the agent is attempting to complete. Truncation, on the other hand, refers to a time-based ending, which occurs when the episode is stopped due to reaching the maximum length, regardless of whether the agent has achieved a terminal state.

\subsection{Value Functions}
In reinforcement learning, we utilize value functions to characterize the quality of a particular state or state-action pair under a specific policy $\pi$ in terms of the expected returns. In other words, value functions provide a way for evaluating a specific policy $\pi$. Specifically, the \emph{\textbf{state-value function}} represents the value of a state, while the \emph{\textbf{action-value function}} reflects the value of an action taken in a given state, or the value of a state-action pair. 

Mathematically, the state-value function for a given state $\bm{s} \in \mathcal{S}$ under a policy $\pi$, denoted $V_\pi(\bm{s})$, represents the expected return when the agent starts in state $\bm{s}$ and subsequently follows policy $\pi$:
\begin{equation}
    V_\pi(\bm{s}) = \mathbb{E}_\pi [G_t \mid \bm{s}_t = \bm{s}] = \mathbb{E}_\pi \left[ \sum_{k=0}^\infty \gamma^k r_{t+k+1} \mid \bm{s}_t = \bm{s} \right]
    \label{eq:def_state_value}
\end{equation}
A higher state-value indicates that the state $\bm{s}_t$ is more favorable under the current policy. For example, under the optimal policy, a state closer to the goal state is more likely to have a higher state-value.

Similarly, we define the action-value function for policy $\pi$, denoted as $Q_\pi(\bm{s}, \bm{a})$, which represents the expected return when taking action $\bm{a}$ in state $\bm{s}$ and following policy $\pi$ thereafter:
\begin{equation}
    Q_\pi(\bm{s}, \bm{a}) = \mathbb{E}_\pi [G_t \mid \bm{s}_t = \bm{s}, \bm{a}_t = \bm{a}] = \mathbb{E}_\pi \left[ \sum_{k=0}^\infty \gamma^k r_{t+k+1} \mid \bm{s}_t = \bm{s}, \bm{a}_t = \bm{a} \right]
    \label{eq:def_action_value}
\end{equation}
The higher the value of a state-action pair, the better the agent selects the action $\bm{a}_t$ in state $\bm{s}_t$ under the current policy. For example, under the optimal policy, for a given state, the action that is more likely to lead the agent toward the goal state is more likely to have a higher action-value.

\begin{figure}[htbp]
    \centering
    \includegraphics[width=\textwidth, trim = 20 95 0 50, clip]{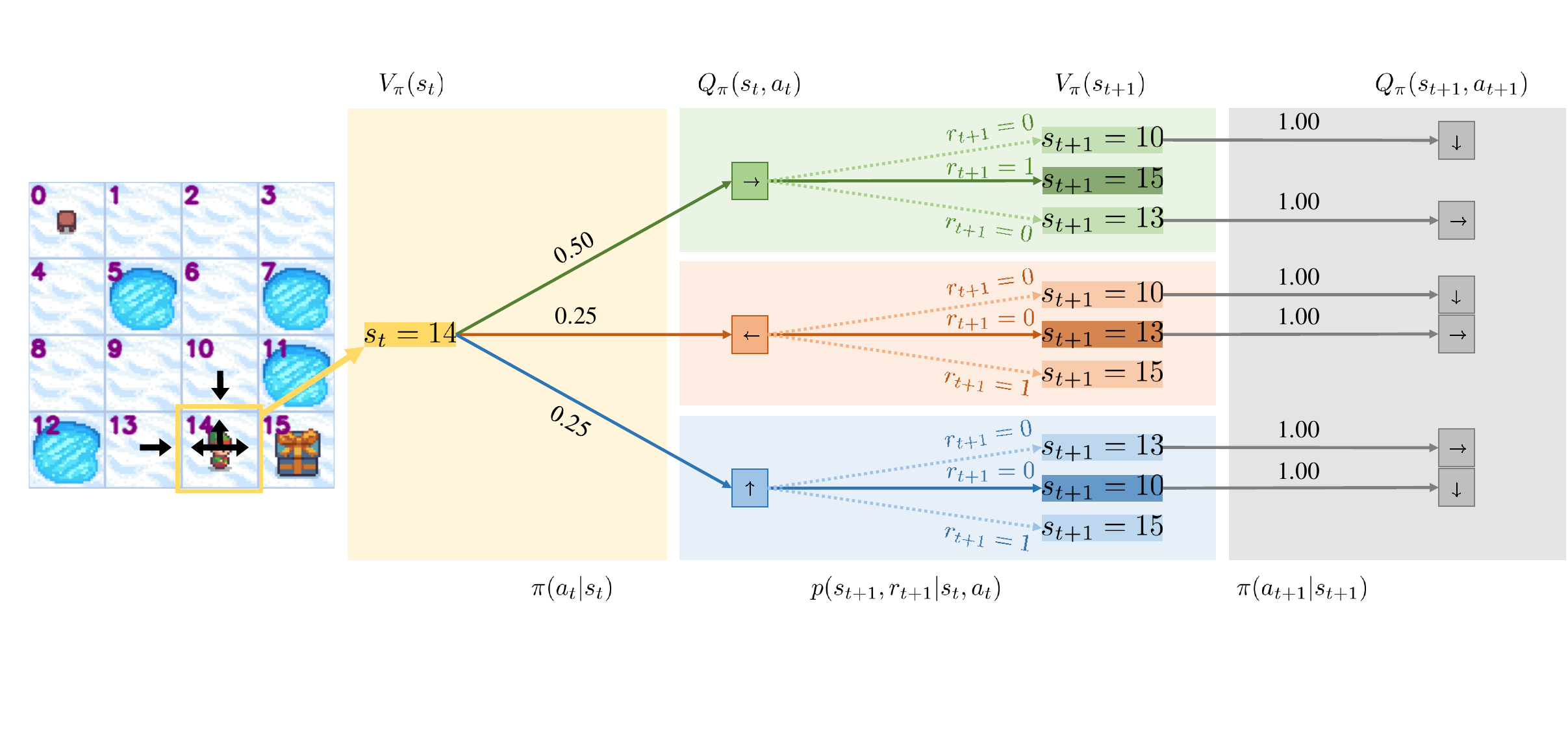}

    \caption{\textbf{An illustration of the relationship between value functions.} This figure depicts a scenario where the lake is slippery. The left panel illustrates a portion of the policy $\pi$ being evaluated. In the highlighted yellow block, the agent is in state 14 and can choose among three actions under policy $\pi$: move right, move left, or move up, with respective probabilities of 0.5, 0.25, and 0.25. The green, orange, and blue blocks represent segments of the potential trajectories following the agent's action. Due to the slipperiness of the lake, the agent may reach the intended next state (solid-line arrows) with a probability of 0.8 or one of two unintended states (dashed-line arrows) with a probability of 0.1 each. The grey block indicates that if the agent reaches the target state, the episode terminates with no future rewards. If the next state is not the target, subsequent actions are deterministic under the current policy $\pi$.}

    \label{fig:relationship_state_action_values}
\end{figure}
\textbf{Relationships between the state-value and action-value:} State-values and action-values are closely interconnected and can be expressed in terms of each other. As illustrated in the yellow block of Figure~\ref{fig:relationship_state_action_values}, the agent is currently in state $s_t = 14$. Under the policy $\pi$, the agent can select one of three possible actions, with probabilities 0.5, 0.25, and 0.25, respectively. Each action corresponds to a unique state-action pair $(s_t, a_t)$, which is associated with a specific action-value $Q_\pi(s_t, a_t)$. The state-value $V_\pi(s_t)$ is the weighted average of the action-values $Q_\pi(s_t, a_t)$, weighted by the probabilities of selecting each action according to the policy, $\pi(a_t | s_t)$. Mathematically, this relationship is given by:
\begin{equation}
    V_\pi(\bm{s}_t) = \sum_{\bm{a}_t} \pi(\bm{a}_t |\bm{s}_t) Q_\pi(\bm{s}_t, \bm{a}_t)
    \label{eq:state_value_from_action_value}
\end{equation}
The green, orange, and blue blocks of Figure~\ref{fig:relationship_state_action_values} illustrate that action-values can also be expressed in terms of state-values. When the agent selects an action $a_t$ in state $s_t$,  it transitions to a next state $s_{t+1}$ with a probability described by the transition model $p(s_{t+1}, r_{t+1} | s_t, a_t)$. Each transition is associated with a reward $r_{t+1}$, and the subsequent trajectory depends on the next state. The action-value $Q_\pi(s_t, a_t)$ is the weighted average of the expected cumulative rewards over all possible subsequent trajectories. This is computed as the immediate reward $r_{t+1}$ plus the discounted state-value of the next state  $\gamma V_\pi(s_{t+1})$. Mathematically, this relationship is given by:
\begin{equation}
    Q_\pi(\bm{s}_t, \bm{a}_t) = \sum_{\bm{s}_{t+1}} \sum_{r_{t+1}} p(\bm{s}_{t+1}, r_{t+1} |\bm{s}_t, \bm{a}_t)\left[ r_{t+1} + \gamma V_\pi(\bm{s}_{t+1}) \right]
    \label{eq:action_value_from_state_value}
\end{equation}

 \textbf{The Bellman equation:} State-values and action-values possess an important recursive property, which gives rise to the Bellman equation. The Bellman equation can be derived by alternately applying the relationships between state-values and action-values, as outlined in Equations~(\ref{eq:state_value_from_action_value}) and~(\ref{eq:action_value_from_state_value}). We begin by introducing the Bellman equation for state-values, which expresses the relationship between the value of a state $s_t$ and the values of its successor states $s_{t+1}$:
\begin{equation}
    \begin{aligned}
        V_\pi(\bm{s}_t) &= \sum_{\bm{a}_t} \pi(\bm{a}_t |\bm{s}_t) {\color{blue}Q_\pi(\bm{s}_t, \bm{a}_t)} & \\
        &= \sum_{\bm{a}_t} \pi(\bm{a}_t |\bm{s}_t) {\color{blue} \sum_{\bm{s}_{t+1}} \sum_{r_{t+1}} p(\bm{s}_{t+1}, r_{t+1} |\bm{s}_t, \bm{a}_t)\left[ r_{t+1} + \gamma V_\pi(\bm{s}_{t+1}) \right]} & { \color{blue}(\text{By Equation~(\ref{eq:action_value_from_state_value})})}
    \end{aligned}
\end{equation}
Similarly, the Bellman equation for action-values expresses the relationship between the value of a state-action pair $(s_t, a_t)$ and the values of its successor state-action pairs $(s_{t+1}, a_{t+1})$:
\begin{equation}
    \begin{aligned}
        Q_\pi(\bm{s}_t, \bm{a}_t) &= \sum_{\bm{s}_{t+1}} \sum_{r_{t+1}} p(\bm{s}_{t+1}, r_{t+1} |\bm{s}_t, \bm{a}_t)\left[ r_{t+1} + \gamma {\color{blue} V_\pi(\bm{s}_{t+1})} \right] & \\
        &= \sum_{\bm{s}_{t+1}} \sum_{r_{t+1}} p(\bm{s}_{t+1}, r_{t+1} |\bm{s}_t, \bm{a}_t)\left[ r_{t+1} + \gamma {\color{blue} \sum_{\bm{a}_{t+1}} \pi(\bm{a}_{t+1} |\bm{s}_{t+1}) Q_\pi(\bm{s}_{t+1}, \bm{a}_{t+1}) } \right] & {\color{blue}(\text{By Equation~(\ref{eq:state_value_from_action_value})})}
    \end{aligned}
\end{equation}

\textbf{Advantage function:} For a given state-action pair $(\bm{s}_t, \bm{a}_t)$, the advantage function $A_\pi(\bm{s}_t, \bm{a}_t)$ is defined as the difference between the action-value of the state-action pair $Q_\pi(\bm{s}_t, \bm{a}_t)$ and the state-value of the state $V_\pi(\bm{s}_t)$:
\begin{equation}
    A_\pi(\bm{s}_t, \bm{a}_t) = Q_\pi(\bm{s}_t, \bm{a}_t) - V_\pi(\bm{s}_t)
\end{equation}
Intuitively, the advantage function measures how much better or worse a particular action $\bm{a}_t$ is compared to the average action in the given state $\bm{s}_t$. Specifically, if $A_\pi(\bm{s}_t, \bm{a}_t) > 0$, selecting action $\bm{a}_t$ yields higher expected returns than the average return obtained by selecting any other action in state $\bm{s}_t$. Conversely, if $A_\pi(\bm{s}_t, \bm{a}_t) < 0$, action $\bm{a}_t$ results in lower expected returns than the average.
\section{The Big Picture of RL Algorithms}
\label{sec:big_picture}
\begin{figure}[htbp]  
    \centering  
    \begin{subfigure}{0.52\textwidth}  
        \centering  
        \includegraphics[width=0.9\linewidth, page=2, trim = 200 90 200 100, clip]{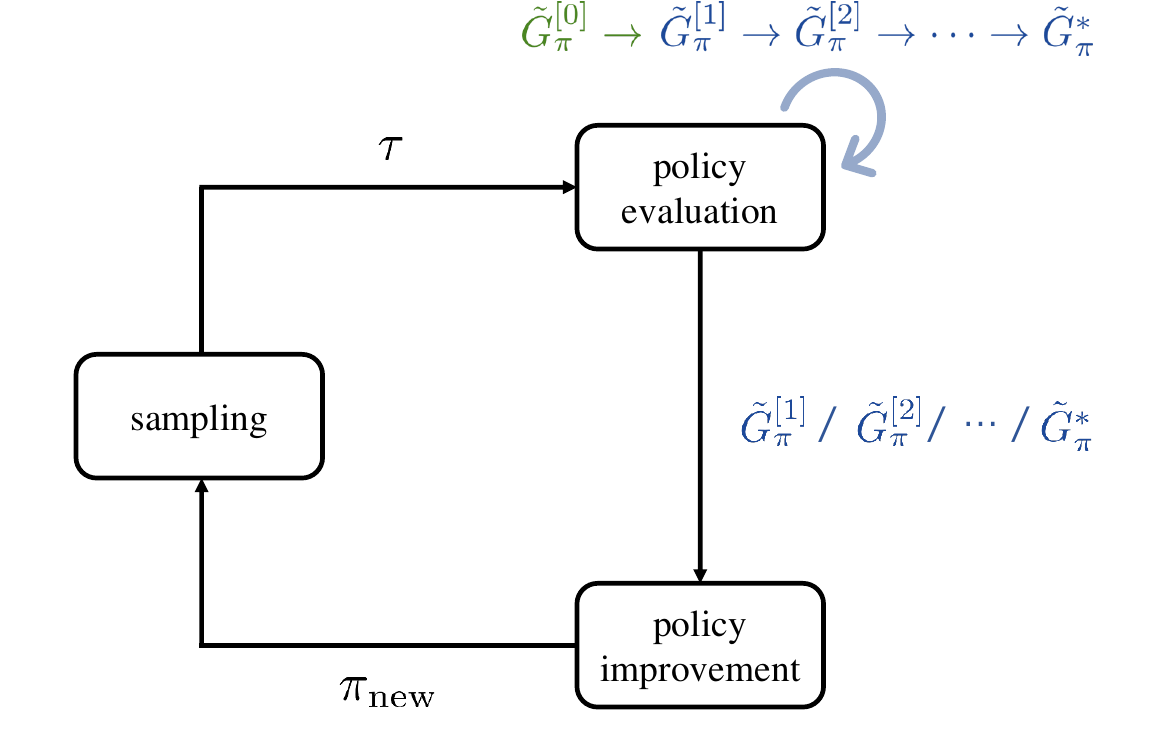}
        \caption{}
        \label{fig:unified_rl_alg_illustration} 
    \end{subfigure}  
    \hfill
    \begin{subfigure}{0.46\textwidth}  
        \centering  
        \includegraphics[width=0.9\linewidth, page=3, trim = 200 90 200 100, clip]{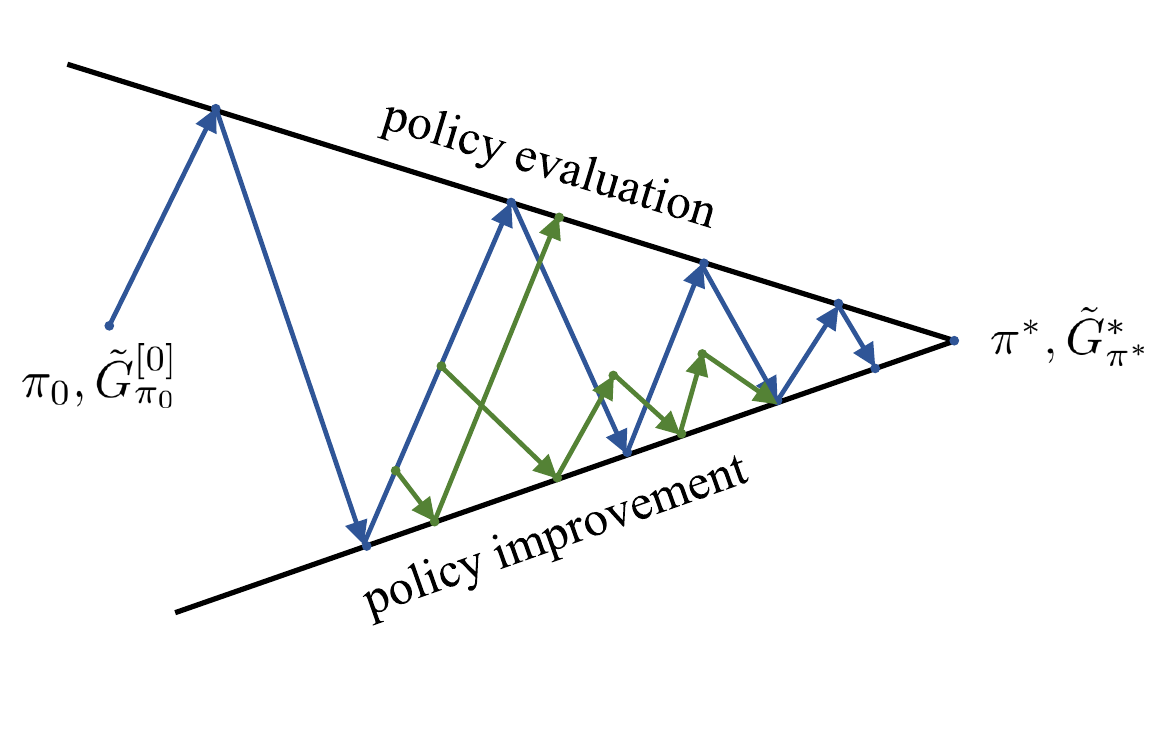}
        \caption{}
        \label{fig:illustration_policy_iteration}
    \end{subfigure}  
    \caption{\textbf{Illustration of Generalized Policy Iteration.} (a) A unified perspective on reinforcement learning algorithms. (b) An illustration of the truncation in the policy evaluation process. The blue arrows indicate the complete policy evaluation process, while the green arrows demonstrate the truncated policy evaluation process.}
\end{figure}
Before delving into the details of various reinforcement learning algorithms, we introduce a unified perspective for analyzing them in this section. 
The objective of reinforcement learning algorithms is to identify the optimal policy, denoted as $\pi^*$, which maximizes the expected accumulated rewards, referred to as the expected return. The expected return is commonly represented using state-values or action-values.  We use $\tilde{G}_\pi$ as an abstract representation of these quantities.
As illustrated in Figure~\ref{fig:unified_rl_alg_illustration}, all the reinforcement learning algorithms discussed in this paper can be framed as \textbf{\textit{generalized policy iterations (GPI)}}, which consists of the following iterative steps:
\begin{enumerate}[nosep]
    \item \textbf{Sampling:} Generate trajectories by allowing the agent to interact with the environment according to the current policy.
    \item \textbf{Policy evaluation:} Assess the goodness of each state or state-action pair under the current policy by calculating the expected return based on the sampled trajectories.
    \item \textbf{Policy improvement:} Refine the policy using the evaluation results to make it greedy with respect to the estimated quality of states or state-action pairs.
\end{enumerate}
The policy evaluation process is said to converge when the value estimates for each state or state-action pair are consistent with the current policy. Similarly, the policy improvement process converges when the policy becomes greedy with respect to the current value estimates. The overall GPI process converges only when a policy $\pi^\star$ is found that is greedy with respect to its evaluation function $\tilde{G}_{\pi^*}^*$.

\begin{figure}[htbp]  
    \centering  
    \begin{subfigure}{0.28\textwidth}  
        \centering  
        \includegraphics[width=\linewidth]{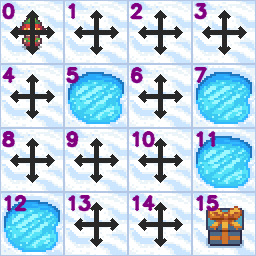}
        \caption{}
        \label{fig:random_policy_illustration}
    \end{subfigure}  
    \hfill
    \begin{subfigure}{0.28\textwidth}  
        \centering
        \includegraphics[width=\linewidth]{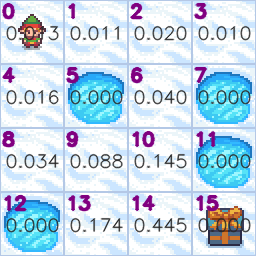}
        \caption{}
        \label{fig:state_values_illustration}
    \end{subfigure}  
    \hfill
    \begin{subfigure}{0.28\textwidth}  
        \centering  
        \includegraphics[width=\linewidth]{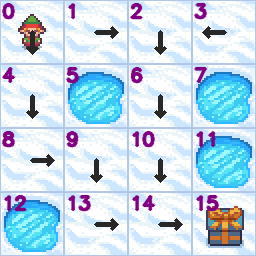}
        \caption{}
        \label{fig:greedy_policy_illustration}
    \end{subfigure}  
    \caption{\textbf{Illustration of an iteration in GPI.} We use the frozen-lake environment as an example.
    (a) The process begins with a random policy, where each action has an equal probability of being selected in all states.
    (b) State values are estimated for each state under the random policy using Monte Carlo methods.
    (c) During the policy improvement procedure, the action leading to the state with the highest state value is selected.}
    \label{fig:policy_evaluation_illustration}
\end{figure}
To clarify this description, we specify $\tilde{G}_\pi$ as the action-values $Q_\pi(\bm{s}_t, \bm{a}_t)$. As introduced in the previous section, $Q_\pi(\bm{s}_t, \bm{a}_t)$ represents the expected return when the agent selects action $\bm{a}_t$ in state $\bm{s}_t$ under the policy $\pi$. It measures the quality of an action taken in a given state under the current policy. Several methods (discussed in the next section) can iteratively estimate the action-values based on sampling. To evaluate a specific policy $\pi$, these methods start with an initial estimate $Q^{[0]}_\pi({\bm{s}_t, \bm{a}_t})$ and iteratively refine it until they converge to the true action-values. During this process, a sequence of estimates is generated for each state-action pair. 
In practice, the true action values are usually unknown, and convergence is determined when the difference between successive estimates becomes sufficiently small:
\begin{equation}
    \max_{\bm{s}_t \in \mathcal{S}, \bm{a}_t \in \mathcal{A}} \quad Q^{[k+1]}_\pi(\bm{s}_t, \bm{a}_t) - Q^{[k]}_\pi(\bm{s}_t, \bm{a}_t) \leq \Delta
\end{equation}
where $Q^{[k]}_\pi(\bm{s}_t, \bm{a}_t)$ denotes the $k$-th estimate for the state-action pair $(\bm{s}_t, \bm{a}_t)$, and $\Delta$ is a predefined tolerance threshold. This condition indicates that further iterations will not significantly alter the value estimates. When convergence is achieved, we denote the final evaluation as $Q^*_\pi(\bm{s}_t, \bm{a}_t)$. Once the policy evaluation process converges, the policy improvement step updates the policy to make it greedy with respect to $Q^*_\pi(\bm{s}_t, \bm{a}_t)$. Specifically, for any state $\bm{s}_t \in \mathcal{S}$, the new policy is defined as:
\begin{equation}
    \pi_\text{new} \leftarrow \mathop{\arg\max}\limits_{\bm{a}_t \in \mathcal{A}} Q_\pi^*(\bm{s}_t, \bm{a}_t)
\end{equation} 
The overall process converges when an optimal policy $\pi^*$ is found that is greedy with respect to its evaluation $Q^*_{\pi^*}(\bm{s}_t, \bm{a}_t)$.

However, the policy evaluation process is often computationally intensive, requiring many iterations to achieve convergence. This raises an important question: must we wait for the policy evaluation to fully converge before performing a policy improvement step? The answer is no. As illustrated in Figure~\ref{fig:illustration_policy_iteration}, it is often sufficient to truncate the policy evaluation process after a limited number of iterations, without compromising the convergence guarantees of the overall GPI process. A notable special case is value iteration, where the policy evaluation is truncated after a single iteration. This approach is widely adopted in many reinforcement learning algorithms.

In the next two sections, we will first focus on the policy evaluation process, introducing several methods to estimate the values of states or state-action pairs. Then, in Section~\ref{sec:policy_gradients}, we shift our focus to the policy improvement process, where we introduce policy gradient methods as an alternative approach to derive the optimal policy based on evaluation.
\section{Value Estimation}
\label{sec:value_estimation}
In this section, we introduce several methods for evaluating a given policy by estimating state-values or action-values through sampling. These methods serve as the foundation for assessing the quality of states or state-action pairs under a specific policy. While the primary focus is on the complete policy evaluation process, which typically requires numerous iterations to achieve convergence, it is important to highlight that these methods can also be truncated to perform a single iteration. Such a truncated evaluation approach is commonly used in various reinforcement learning algorithms to accelerate the iterative process of Generalized Policy Iteration (GPI).

The following subsections provide a detailed discussion of these methods. First, we explore Monte Carlo methods, a fundamental approach that relies on complete trajectories for estimation. Next, we introduce Temporal-Difference (TD) methods, which offer the advantage of not requiring full trajectories. Finally, we delve into the TD($\lambda$) framework, a generalized method that bridges Monte Carlo and TD approaches, enabling a trade-off between variance and bias in the estimation process.

\subsection{Monte-Carlo (MC) Methods}

In reinforcement learning, \textbf{\textit{Monte-Carlo (MC) methods}} offer a simple yet effective approach for evaluating a given policy. The core idea is to compute the average of observed returns after visiting each state. By the law of large numbers, as more episodes are sampled, the average of these returns converges to their true values under the given policy. Specifically, when estimating the state-value $V_\pi(\bm{s}_i)$ for each state $\bm{s}_i$, we maintain a list $\mathcal{G}(\bm{s}_i)$ for each state, which stores the observed accumulated rewards obtained when visiting that state. The final estimation of the state-value $V_\pi(\bm{s}_i)$ under policy $\pi$ is the average of all observations in $\mathcal{G}(\bm{s}_i)$. This procedure is formally described in Algorithm~\ref{alg:every_visit_mc_state_value}. 
\IncMargin{1.2em}

\begin{algorithm}[h]
    \caption{Policy Evaluation Using Monte-Carlo Method}
    \label{alg:every_visit_mc_state_value}
    \LinesNumbered
    
    \KwIn {the policy $\pi$ to be evaluated, discounted factor $\gamma$, number episodes $N$}
    \KwOut {The value of each state $V_\pi(\bm{s})$ under policy $\pi$}

    \textbf{Initialization: } For all $\bm{s}_i \in \mathcal{S}$, initialize an empty list $\mathcal{G}(\bm{s}_i)$. \

    \ForEach{sampling episode $k \in [1, N]$}
    {
        Set the initial state $\bm{s}_0 \in \mathcal{S}$ of the agent \
        
        Generate a trajectory $\tau = (\bm{s}_0, \bm{a}_0, r_1, \cdots, \bm{s}_{T-1}, \bm{a}_{T-1}, r_T)$ by letting the agent interact with the environment using policy $\pi$\

        $G_t \leftarrow 0$\

        \ForEach{$t = T-1, T-2, \cdots, 0$}
        {
            $G_t \leftarrow \lambda G_t + r_{t+1}$ \
        
            Find the list $\mathcal{G}(\bm{s}_t)$ corresponding to $\bm{s}_t$, and add $G_t$ to this list.\

            Updating the estimate of $V_\pi(\bm{s}_t)$: computing the average of all values in the list $\mathcal{G}(\bm{s}_t)$
        }
    }
\end{algorithm}

\DecMargin{1.2em}

\textbf{Incremental implementation:} In Algorithm~\ref{alg:every_visit_mc_state_value}, we need to store every observation in $\mathcal{G}(\bm{s}_i)$. If the number of samples is very large, this approach can be space-intensive. To address this issue, we can use an incremental mean calculation method:
\begin{subequations}
    \begin{align}
        \hat{V}^{[m+1]}_\pi(\bm{s}_t) &= \frac{1}{m+1} \sum_{i=1}^{m+1} G_t^{[i]} \label{eq:nonincremental_mc_v}\\
        &= \frac{1}{m+1} \left( \sum_{i=1}^{m} G_t^{[i]} + G_t^{[m+1]} \right) \notag \\
        &= \frac{1}{m+1} {\color{blue} \cdot m \cdot \frac{1}{m}} \sum_{i=1}^{m} G_t^{[i]} + \frac{1}{m+1} G_t^{[m+1]} \notag \\
        &= \frac{1}{m+1} \cdot m \hat{V}_\pi^{[m]}(\bm{s}_t) + \frac{1}{m+1} G_t^{[m+1]} \notag \\
        &= \frac{1}{m+1} \left( (m+1) \hat{V}_\pi^{[m]}(\bm{s}_t) - \hat{V}_\pi^{[m]}(\bm{s}_t) + G_t^{[m+1]} \right) \notag \\
        &= \hat{V}_\pi^{[m]}(\bm{s}_t) + \frac{1}{m+1} \left( G_t^{[m+1]} - \hat{V}_\pi^{[m]}(\bm{s}_t) \right),
        \label{eq:incremental_mc_v}        
    \end{align}
\end{subequations}
where $G_t^{[m]}$ represents the observed return the $m$-th time we visit state $\bm{s}_t$, and $\hat{V}_\pi^{[m]}(\bm{s}_t)$ denotes the estimate of the state-value function $V_\pi(\bm{s}_t)$ after $m$ updates. Equation~(\ref{eq:incremental_mc_v}) can be written in a more general form:
\begin{equation}
    \text{New Estimation} \leftarrow \text{Old Estimation} + \alpha \left( \text{New Observation} - \text{Old Estimation} \right),
    \label{eq:incremental_general_form}
\end{equation}
where $\alpha = \frac{1}{m+1}$. Generally, Equation~(\ref{eq:incremental_general_form}) allows us to adjust the previous estimation by a fraction of the difference between the new observation and the old estimation, with $\alpha$ controlling the speed of the update. Using this incremental manner, we only need to store the previous estimation and the current observation, which significantly reduces the memory requirements and improves the computational efficiency of Algorithm~\ref{alg:every_visit_mc_state_value}.


\textbf{MC methods for estimating action-values:} Given a policy $\pi$ to evaluate, the procedure for estimating action-values using the Monte Carlo method is essentially the same as that for estimating state-values. The only difference is that we need to maintain a list for each state-action pair, rather than for each state, to store observations. Similarly, action-value estimates can also be updated incrementally:
\begin{equation}
    \hat{Q}_\pi^{[m+1]}(\bm{s}_t, \bm{a}_t) = \hat{Q}_\pi^{[m]}(\bm{s}_t, \bm{a}_t) + \alpha \left(G_t^{[m+1]} - \hat{Q}_\pi^{[m]}(\bm{s}_t, \bm{a}_t)\right),
\end{equation}
where $G_t^{[m]}$ represents the observed return the $m$-th time we visit state-action pair ($\bm{s}_t, \bm{a}_t$), and $\hat{Q}_\pi^{[m]}(\bm{s}_t, \bm{a}_t)$ denotes the estimate of the action-value function $Q_\pi(\bm{s}_t, \bm{a}_t)$ after $m$ updates. 

\subsection{Temporal-Difference (TD) Methods}
\label{sec:td_methods}
\begin{figure}[htbp]
    \centering
    \includegraphics[width=\textwidth, trim = 0 80 0 25, clip]{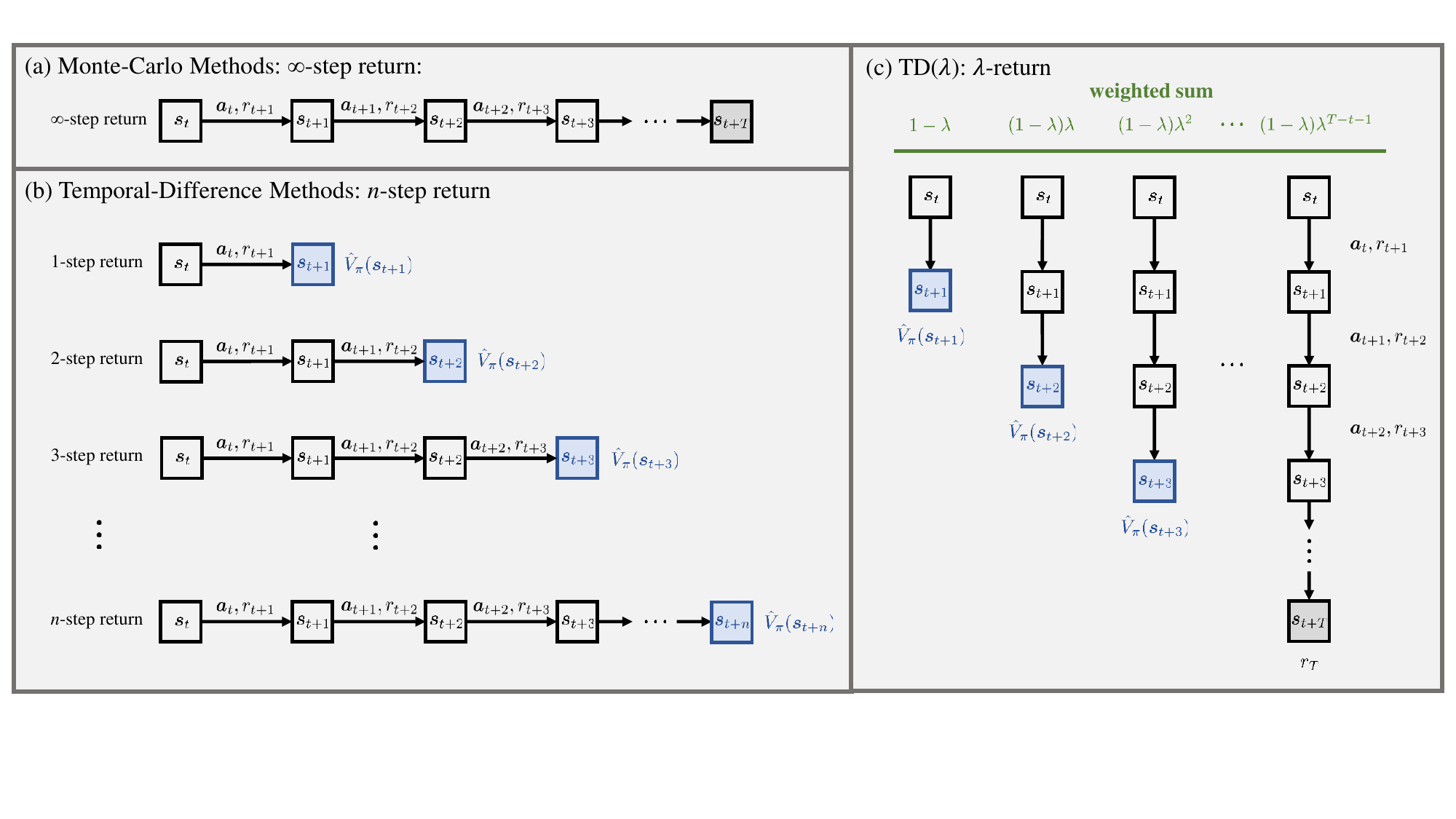}

    \caption{\textbf{Illustration of different methods for estimating the state-value of a state.} The unfilled square nodes represent the use of observed rewards after visiting a state, the blue square nodes indicate the use of values from a previous estimate, and the gray-shaded square nodes represent terminal states in the trajectory. (a) Monte Carlo (MC) methods require a complete trajectory to calculate returns. (b) Temporal-Difference (TD) methods eliminate the need for the entire trajectory by using a previous estimate of $V_\pi(\bm{s}_{t+n})$. (c) TD($\lambda$) utilizes the $\lambda$-return to estimate state-values, where the $\lambda$-return is calculated as an exponentially-weighted average of $n$-step returns, with decay parameter $\lambda$, along with the normalization factor $1 - \lambda$.}

    \label{fig:mc_td_illustration}
\end{figure}

While Monte Carlo (MC) methods can effectively estimate values, they have a significant limitation: updates occur only at the end of an episode. In applications with very long episodes, delaying updates until completion can be inefficient. Moreover, some applications involve episodes of infinite length, making MC methods impractical for value updates. MC methods are also unsuitable for online learning. To address these limitations, \textbf{\textit{Temporal-Difference (TD)}} methods have been introduced. Like MC methods, TD methods build upon Equation~(\ref{eq:incremental_general_form}), but unlike MC, TD methods update values at each time step. This section provides a comprehensive introduction to TD methods.

In MC methods, updates must wait until the end of an episode because calculating the return requires summing all rewards obtained after visiting a particular state. To avoid this delay, alternative methods for calculating returns are needed. Specifically, when evaluating a given policy $\pi$ via state-values, we consider the computation of the return from a trajectory $ \tau = \{\bm{s}_0, \bm{a}_0, r_1, \dots, \bm{s}_{T-1}, \bm{a}_{T-1}, r_T, \bm{s}_T\} $. The return for an agent starting in state $ \bm{s}_t $ is defined by Equation~(\ref{eq:Gt_original}). Notably, the term $ r_{t+2} + \cdots + \gamma^{T-t-2}r_T $ represents the return for the agent starting in state $ \bm{s}_{t+1} $. Based on this observation, we can express the return $ G_t $ as $ G_t = r_{t+1} + \gamma G_{t+1} $. As shown in Equation~(\ref{eq:Gt_approx}), TD methods approximate $ G_{t+1} $ using the previous estimate of $ V_\pi (\bm{s}_{t+1}) $, denoted as $ \hat{V}_\pi (\bm{s}_{t+1}) $, thereby avoiding the need to compute $G_{t+1}$ through the sequence $r_{t+2}, \cdots, r_T$. Consequently, this approach elininates the need to wait until the end of an episode for updates.
\begin{subequations}
    \begin{align}
        G_t &= r_{t+1} + \underbrace{\gamma r_{t+2} + \dots + \gamma^{T-t-1}r_T}_{\gamma G_{t+1}} \label{eq:Gt_original} \\
        &\approx r_{t+1} + \gamma \hat{V}_\pi(\bm{s}_{t+1}) \label{eq:Gt_approx}
    \end{align}
\end{subequations}
We can see that the TD method allows us to compute a new estimate of $V_\pi(\bm{s}_t)$ by observing only $ r_{t+1} $, i.e., the reward at the next time step. By substituting Equation~(\ref{eq:Gt_approx}) into Equation~(\ref{eq:incremental_mc_v}), we derive the TD update rule for estimating state values:
\begin{equation}
    \hat{V}^{[m+1]}_\pi(\bm{s}_t) = \hat{V}_\pi^{[m]}(\bm{s}_t) + \frac{1}{m+1} \left( r_{t+1} + \gamma \hat{V}_\pi(\bm{s}_{t+1}) - \hat{V}_\pi^{[m]}(\bm{s}_t) \right),
\end{equation}
where $ r_{t+1} + \gamma \hat{V}_\pi(\bm{s}_{t+1}) $ is referred to as the \textbf{\textit{TD target}}, and $ r_{t+1} + \gamma \hat{V}_\pi(\bm{s}_{t+1}) - \hat{V}_\pi^{[m]}(\bm{s}_t) $ is known as the \textbf{\textit{TD error}}.

\textbf{Using $n$-step return:} 
Equation~(\ref{eq:Gt_approx}) is also known as the \textbf{\textit{1-step return}} because it approximates $G_t$ using a linear combination of the reward over only the next 1 step and the estimated state-value of $\bm{s}_{t+1}$. This approach can be generalized to the \textbf{\textit{$n$-step return}}, where we approximate $G_t$ using the linear combination of rewards over the next $n$ steps and the estimated state-value of $\bm{s}_{t+n}$: 
\begin{subequations}
    \begin{align}
        1\text{-step TD:} \quad \quad G_t \approx r_{t+1} + &\gamma \hat{V}_\pi (\bm{s}_{t+1}) \\
        2\text{-step TD:} \quad \quad G_t \approx r_{t+1} + &\gamma r_{t+2} + \gamma^2 \hat{V}_\pi (\bm{s}_{t+2}) \\
        3\text{-step TD:} \quad \quad G_t \approx r_{t+1} + &\gamma r_{t+2} + \gamma^2 r_{t+3} + \gamma^3 \hat{V}_\pi (\bm{s}_{t+3}) \\
        &\vdots \notag \\
        n\text{-step TD:} \quad \quad G_t \approx r_{t+1} + &\cdots + \gamma^{n-1} r_{t+n} + \gamma^n \hat{V}_\pi (\bm{s}_{t+n}) \label{eq:n_step_return_v}
    \end{align}
\end{subequations}
TD methods that use $n$-step returns to approximate the new observation in Equation~(\ref{eq:incremental_general_form}) are referred to as \textbf{\textit{$n$-step TD methods}}. These methods form a spectrum, with 1-step TD methods at one end and Monte Carlo (MC) methods at the other. In this context, MC methods could be considered as $\infty$-step TD methods.

\textbf{TD methods for estimating action-values:} Similar as we did in MC methods, the TD update rule for action-values is as follows:
\begin{equation}
     \hat{Q}_\pi^{[m+1]}(\bm{s}_t, \bm{a}_t) = \hat{Q}_\pi^{[m]}(\bm{s}_t, \bm{a}_t) + \alpha \left(r_{t+1}^{[m+1]} + \gamma \hat{Q}_\pi(\bm{s}_{t+1}, \bm{a}_{t+1}) - \hat{Q}_\pi^{[m]}(\bm{s}_t, \bm{a}_t)\right),
\end{equation}
where $\hat{Q}_\pi(\bm{s}_{t}, \bm{a}_{t})$ denotes an estimate of $Q_\pi(\bm{s}_{t}, \bm{a}_{t})$. We can also leverage the $n$-step return to approximate the new observation in Equation~(\ref{eq:incremental_general_form}):
\begin{equation}
    \begin{aligned}
        G_t &= r_{t+1} + \cdots + \gamma^{n-1} r_{t+n} + \underbrace{\gamma^n r_{t+n+1} + \cdots + \gamma^{T-t-1}r_T}_{\gamma^n G_{t+n}} \\
        &\approx r_{t+1} + \cdots + \gamma^{n-1} r_{t+n} + \gamma^n \hat{Q}_\pi(\bm{s}_{t+n}, \bm{a}_{t+n})
    \end{aligned}
    \label{eq:n_step_return_q}
\end{equation}

\subsection{Trade-off Between Bias and Variance}
\label{sec:td_lambda}
Previously, we have introduced two approaches, MC and TD methods, for policy evaluation by estimating value functions under a given policy. Generally, MC methods provide unbiased but relatively high-variance estimates, while TD methods offer lower variance at the cost of increased bias. Specifically, MC methods yield an unbiased estimate of the expected return for a given state by calculating returns as the sum of all future rewards. However, this approach results in high variance, as it depends on the cumulative sum of numerous random variables $ r_{t+1}, r_{t+2}, \dots, r_T $, each affected by the policy’s action choices and the environment's state-transition dynamics. In contrast, the 1-step TD method reduces variance by involving only two random variables ($r_{t+1}$ and $\bm{s}_{t+1}$) for state-values estimation, or three random variables ($r_{t+1}$, $\bm{s}_{t+1}$, and $\bm{a}_{t+1}$) for action-values estimation.
Despite this advantage, TD methods introduce bias due to the use of the biased estimator—specifically, $ \hat{V}_\pi(\bm{s}_{t+1}) $ for state-value estimation and $ \hat{Q}_\pi(\bm{s}_{t+1}, \bm{a}_{t+1}) $ for action-value estimation—in place of the full return. In TD methods that use $ n $-step returns, as shown in Equations~(\ref{eq:n_step_return_v})(\ref{eq:n_step_return_q}), the choice of $ n $ serves as a trade-off between bias and variance in the value estimation. In this section, we introduce the \textbf{\textit{$\lambda$-return}}, which provides another method to trade off between the bias and variance of the estimator. Estimating the value functions using the TD method computed with the $\lambda$-return results in the \textbf{\textit{TD($\lambda$)}} algorithm~\cite{rl_an_intro_sutton_2018}. Empirical studies suggest that the $\lambda$-return can achieve better performance than using a fixed $ n $-step return~\cite{rl_an_intro_sutton_2018, gae_iclr2016}.

The $\lambda$-return, denoted by $G_t^\lambda$, is calculated as an exponentially-weighted average of $n$-step returns with a decay parameter $\lambda \in [0, 1]$, along with the normalization factor $1 - \lambda$ to ensure the sum of all weights equals 1.
\begin{equation}
    \begin{aligned}
        G_t^\lambda &= (1 - \lambda) \sum_{l=1}^\infty \lambda^{l-1} G_{t:t+l} 
        \quad \quad {\color{blue} \left(\text{sum of weights: }(1 - \lambda)\sum_{l=1}^\infty \lambda^{l-1} = (1 - \lambda) \frac{1}{1 - \lambda} = 1 \right)} \\
        &= (1-\lambda) \left( G_{t:t+1} + \lambda G_{t:t+2} + \lambda^2 G_{t:t+3} + \cdots \right),
    \end{aligned}
    \label{eq:TD_lambda_def}
\end{equation}
where $G_{t:t+n}$ represents the $n$-step return. Simplifying Equation~(\ref{eq:TD_lambda_def}) yields:
\begin{equation}
    \begin{aligned}
        G_t^\lambda &= (1-\lambda) \left( G_{t:t+1} + \lambda G_{t:t+2} + \lambda^2 G_{t:t+3} + \cdots \right) \\
        &= (1-\lambda)\left( r_{t+1} + \gamma \hat{V}_\pi (\bm{s}_{t+1}) \right) + (1-\lambda) \lambda \left( r_{t+1} + \gamma r_{t+2} + \gamma^2 \hat{V}_\pi (\bm{s}_{t+2}) \right) \\
        & \quad \quad + (1-\lambda) \lambda^2 \left( r_{t+1} + \gamma r_{t+2} + \gamma^2 r_{t+3} + \gamma^3 \hat{V}_\pi (\bm{s}_{t+3}) \right) + \cdots \\
        &= (1 - \lambda) \left[ (1 + \lambda + \lambda^2 + \cdots) r_{t+1} + \gamma \lambda (1 + \lambda + \lambda^2 + \cdots) r_{t+2} + \gamma^2 \lambda^2 (1 + \lambda + \lambda^2 + \cdots) r_{t+3} + \cdots \right] \\
        & \quad \quad + (1-\lambda)\left( \gamma \hat{V}_\pi(\bm{s}_{t+1}) + \gamma^2 \lambda \hat{V}_\pi(\bm{s}_{t+2}) + \gamma^3 \lambda^2 \hat{V}_\pi(\bm{s}_{t+3}) + \cdots \right) \\
        &= (1 - \lambda) \left[ \left(\frac{1}{1 - \lambda}\right) r_{t+1} + \gamma \lambda \left(\frac{1}{1 - \lambda}\right) r_{t+2} + \gamma^2 \lambda^2 \left(\frac{1}{1 - \lambda}\right) r_{t+3} + \cdots \right] \\
        & \quad \quad + (1-\lambda)\left( \gamma \hat{V}_\pi(\bm{s}_{t+1}) + \gamma^2 \lambda \hat{V}_\pi(\bm{s}_{t+2}) + \gamma^3 \lambda^2 \hat{V}_\pi(\bm{s}_{t+3}) + \cdots \right)\\
        &= \sum_{l=1}^\infty (\gamma \lambda)^{l - 1} r_{t+l} + \gamma^l \lambda^{l-1} (1 - \lambda) \hat{V}_\pi (\bm{s}_{t+l})
    \end{aligned}
\end{equation}
where we observe two special cases by setting $\lambda = 0$ and $\lambda = 1$:
\begin{subequations}
    \begin{align}
        \lambda &= 0: \quad G_t^0 = r_{t+1} + \gamma \hat{V}_\pi (\bm{s}_{t+1}) \quad \quad {\color{blue} (0^0 = 1, 0^1 = 0, 0^2 = 0, \cdots)} \\
        \lambda &= 1: \quad G_t^1 = \sum_{l=1}^\infty \gamma^{l-1} r_{t+l} = r_{t+1} + \gamma r_{t+2} + \gamma^2 r_{t+3} + \cdots
    \end{align}
\end{subequations}
It is evident that when $\lambda = 1$, estimating values using $\lambda$-return leads to the Monte Carlo methods, while $\lambda$-return reduces to the 1-step return when $\lambda = 0$. For values $0 < \lambda < 1$, the $\lambda$-return offers a compromise between bias and variance, controlled by the parameter $\lambda$.

Now we consider episodes with a finite length $T$. Corresponding to the previous infinite-horizon case, we can assume that all rewards after step $T$ are 0 so that we have $G_{t:t+n} = G_{t:T}$ for all $n \geq T - t$. Consequently, we have
\begin{equation}
    \begin{aligned}
        G_t^\lambda &= (1 - \lambda) \sum_{l=1}^\infty \lambda^{l-1} G_{t:t+l} \\
        &= (1 - \lambda) \sum_{l=1}^{T-t-1} \lambda^{l-1} G_{t:t+l} + (1 - \lambda) \sum_{l=T-t}^\infty \lambda^{l-1}G_{t:T} \\
        &= (1 - \lambda) \sum_{l=1}^{T-t-1} \lambda^{l-1} G_{t:t+l} + (1 - \lambda) \lambda^{T-t-1}(1 + \lambda + \lambda^2 + \cdots)G_{t:T} \\
        &= (1 - \lambda) \sum_{l=1}^{T-t-1} \lambda^{l-1} G_{t:t+l} + \lambda^{T-t-1}G_{t:T}
    \end{aligned}
\end{equation}
The 1-step return is given the largest weight, $1 - \lambda$; the 2-step return is given the next largest weight; and so on. 

\subsection{Value Parameterization}
\label{sec:value_parameterization}
\begin{figure}[htbp]
    \centering
    \includegraphics[width=0.9\textwidth, trim = 0 200 0 165, clip]{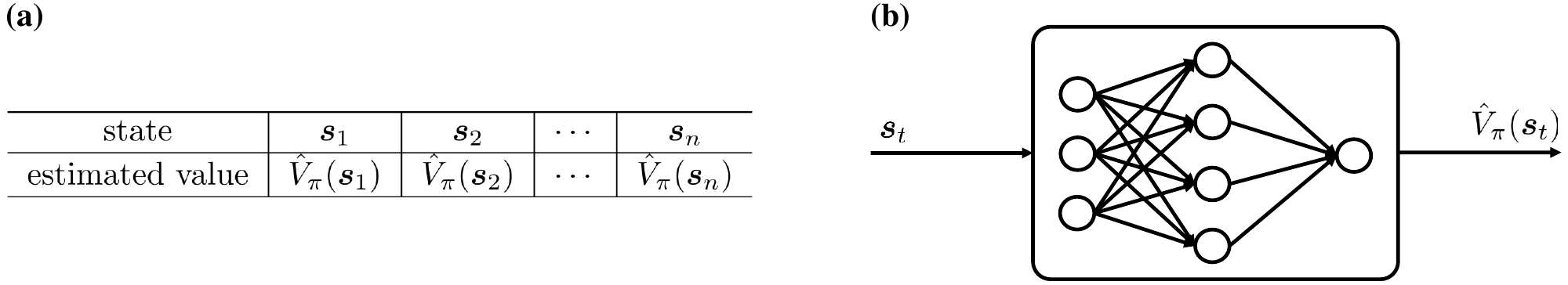}

    \caption{\textbf{Illustration of different methods for storing estimated values.} (a) Tabular storage. (b) Function approximator, with deep neural networks being a common choice.}

    \label{fig:value_parameterization}
\end{figure}
The methods introduced previously are referred to as \textbf{\textit{tabular-based methods}}. These methods are not practical for high-dimensional cases, where the state or action spaces are extremely large, making tabular storage infeasible. To address this limitation, we can use a function approximator $f(\cdot; \bm{w})$ to approximate the value function. In this context, the value functions are parameterized by $\bm{w}$. For instance, when parameterizing the state-value functions, the function approximator takes a state $\bm{s}$ as input and outputs the corresponding state-value under a specific policy $\pi$, i.e. $f(\cdot; \bm{w}) = \hat{V}_\pi(\bm{s}; \bm{w})$.

A commonly used function approximator is a deep neural network.  The goal is to optimize the parameters $\bm{w}$ to achieve the best approximation of the true value funtions. Specifically, for approximating the state-value function under a given policy $\pi$. We aim to minimize the difference between the output of the state-value function approximator and the true state-value $y_t$ for any given state $\bm{s}_t \in \mathcal{S}$:
\begin{equation}
    \bm{w}^* = \mathop{\arg\min}\limits_{\bm{w}} \mathbb{E}_{\pi}\left[ \left( y_t - \hat{V}_\pi(\bm{s}_t; \bm{w}) \right)^2 \right]
\end{equation}

In practical applications, the true state-value is not directly accessible. Instead, we use methods introduced in previous sections to define the target $y_t$. One common choice is the TD target. The objective with the TD target becomes
\begin{equation}
    \bm{w}^* = \mathop{\arg\min}\limits_{\bm{w}} \mathbb{E}_{(\bm{s}_t, r_t, \bm{s}_{t+1}) \sim \pi}\left[ \left( r_t + \gamma \hat{V}_\pi(\bm{s}_{t+1}; \bm{w}) - \hat{V}_\pi(\bm{s}_t; \bm{w}) \right)^2 \right]
\end{equation}

From the perspective of deep learning, this involves training a neural network $\hat{V}_\pi(\bm{s}; \bm{w})$ within a supervised learning framework, where the task is framed as a regression problem. The training data is generated by having the agent interact with the environment following the policy $\pi$. Specifically, suppose the current network parameters are $\bm{w}_c$, a sampled trajectory $\tau = \{\bm{s}_t, \bm{a}_t, r_{t+1}\}_{t=0}^{T-1}$ generates training pairs $(r_{t+1} + \gamma \hat{V}_\pi(\bm{s}_{t+1}; \bm{w}_c), \hat{V}_\pi(\bm{s}_t; \bm{w}_c))$. Using this dataset, we train the neural network by minimizing the mean squared error (MSE) loss:
\begin{equation}
    \mathcal{L}(\bm{w}) = \frac{1}{N} \sum_i \left( r_{t+1}^{(i)} + \gamma \hat{V}_\pi(\bm{s}_{t+1}^{(i)}; \bm{w}) - \hat{V}_\pi (\bm{s}_t^{(i)}; \bm{w}) \right)^2
\end{equation}
where $N$ denotes the total number of training pairs, and $\bm{s}_t^{(i)}$ and $r_{t+1}^{(i)}$ represent the $i$-th occurrence of the state $\bm{s}_t$ and the corresponding rewards in the samples, respectively.




\section{Policy Gradients}
\label{sec:policy_gradients}
In the previous sections, we discussed various methods for evaluating a specific policy. A straightforward way to improve a policy is by selecting actions greedily based on these evaluation results. In this section, we focus on policy gradient methods, another fundamental and widely-used approach for policy improvement.

Recall that the goal of reinforcement learning is to identify the optimal policy that maximizes the expected return $G(\tau)$. Examples of $G(\tau)$ include the following~\cite{gae_iclr2016}:
\begin{itemize}[nosep]
    \item $\sum_{t=1}^T r_t$: the total reward of the trajectory.
    \item $\sum_{t=t^\prime+1}^T r_t$: the reward accumulated after taking action $\bm{a}_{t^\prime}$.
    \item $Q_\pi(\bm{s}_t, \bm{a}_t)$: the action-value function.
    \item $A_\pi(\bm{s}_t, \bm{a}_t)$: the advantage function.
\end{itemize}

Similar to value parameterization, we can parameterize the policy $\pi$. A common approach is to use a deep neural network, denoted as $\pi(\bm{a} | \bm{s}; \bm{\theta})$ or $\pi_{\bm{\theta}}$, where $\bm{\theta}$ represents the network's weights. This network maps states $\bm{s} \in \mathcal{S}$ to actions $\bm{a} \in \mathcal{A}$. In this case, $\pi(\bm{a} | \bm{s}; \bm{\theta})$ is differentiable.

Evaluating a policy $\pi_{\bm{\theta}}$ involves computing the expected return $\mathbb{E}_{\pi_{\bm{\theta}}}[G(\tau)]$ under the current policy parameters $\bm{\theta}$. The gradient of $\mathbb{E}_{\pi_{\bm{\theta}}}[G(\tau)]$ indicates the direction of the steepest ascent in the expected return with respect to the policy parameters $\bm{\theta}$. To maximize the expected return and improve the policy, a gradient ascent step is performed to update the policy parameters:
\begin{equation}
    \bm{\theta}_{\text{new}} \leftarrow \bm{\theta}_{\text{old}} + \alpha \nabla_{\bm{\theta}_\text{old}} \mathbb{E}_{\pi_{\bm{\theta}_\text{old}}}[G(\tau)],
    \label{eq:pg_param_update}
\end{equation}
where $\alpha$ is the learning rate.  Algorithms that follow this framework are known as \textbf{\textit{policy gradient methods}}.

In the subsequent subsections, we first introduce the basic algorithm based on this framework, which is referred to as the vanilla policy gradient (VPG) algorithm. Next, we discuss the integration of value parameterization, which gives rise to actor-critic methods. Additionally, we explore techniques for leveraging off-policy samples to improve data efficiency. Finally, we present the proximal policy optimization (PPO) algorithm, one of the most popular deep reinforcement learning methods in practical applications.

\subsection{REINFORCE: Monte-Carlo Policy Gradient}
The primary objective of this subsection is to derive a practical gradient estimator for $\nabla_{\bm{\theta}_\text{old}} \mathbb{E}_{\pi_{\bm{\theta}_\text{old}}}[G(\tau)]$. Additionally, we introduce variance reduction techniques to enhance the overall performance and stability of the algorithm. The resulting method is commonly referred to as the \textbf{\textit{Vanilla Policy Gradient (VPG)}} algorithm.

We begin our discussion be treating the trajectory $\tau = \{\bm{s}_t, \bm{a}_t, r_{t+1}\}_{t=0}^{T-1}$ as a random variable, with $p_{\bm{\theta}}(\tau)$ representing the probability of observing a specific trajectory under the policy parameterized by $\bm{\theta}$.
We begin our discussion by considering the undiscounted case where the total accumulated reward is given by $G(\tau) = \sum_{t=1}^T r_t$. Our goal is to optimize the policy parameters $\bm{\theta}$ to maximize the expected return. To achieve this, we focus on computing the gradient $\nabla_{\bm{\theta}} \mathbb{E}_{\tau \sim p_{\bm{\theta}}(\tau)}[G(\tau)]$:
\begin{equation}
    \begin{aligned}
        \nabla_{\bm{\theta}} \mathbb{E}_{\tau \sim p_{\bm{\theta}}(\tau)}[G(\tau)] &= \nabla_{\bm{\theta}} \int p_{\bm{\theta}}(\tau) G(\tau) \text{d}\tau \\
        &= \int \nabla_{\bm{\theta}} p_{\bm{\theta}}(\tau) G(\tau) \text{d}\tau \\
        &= \int {\color{blue} p_{\bm{\theta}} (\tau)} \frac{\nabla_{\bm{\theta}} p_{\bm{\theta}}(\tau)}{{\color{blue} p_{\bm{\theta}}(\tau)}} G(\tau) \text{d}\tau \\
        &= \int p_{\bm{\theta}}(\tau) \nabla_{\bm{\theta}} \log p_{\bm{\theta}} (\tau) G(\tau) \text{d}\tau \\
        &= \mathbb{E}_{\tau \sim p_{\bm{\theta}}(\tau)} \left[ \nabla_{\bm{\theta}} \log p_{\bm{\theta}}(\tau) G(\tau) \right]
    \end{aligned}
    \label{eq:pg_gradient_v1}
\end{equation}
Next, we expand the $\nabla_{\bm{\theta}} \log p_{\bm{\theta}} (\tau)$ to derive a practical formula:
\begin{equation}
    \begin{aligned}
        \nabla_{\bm{\theta}} \log p_{\bm{\theta}}(\tau) &= \nabla_{\bm{\theta}} \log \left[ p({\bm{s}_0}) \pi(\bm{a}_0|\bm{s}_0; \bm{\theta}) p(\bm{s}_1, r_1 | \bm{s}_0, \bm{a}_0) \cdots \pi(\bm{a}_{T-1}|\bm{s}_{T-1}; \bm{\theta}) p(\bm{s}_T, r_T | \bm{s}_{T-1}, \bm{a}_{T-1}) \right] \\
        &= \nabla_{\bm{\theta}} [ \log p({\bm{s}_0}) + \log \pi(\bm{a}_0|\bm{s}_0; \bm{\theta}) + \log p(\bm{s}_1, r_1 | \bm{s}_0, \bm{a}_0) + \cdots + \log \pi(\bm{a}_{T-1}|\bm{s}_{T-1}; \bm{\theta}) \\
        & \quad \quad + \log p(\bm{s}_T, r_T | \bm{s}_{T-1}, \bm{a}_{T-1}) ] \quad \quad \quad {\color{blue} (\log(ab) = \log a + \log b)} \\
        &= \nabla_{\bm{\theta}} \sum_{t=0}^{T-1} \log \pi(\bm{a}_t|\bm{s}_t; \bm{\theta})
    \end{aligned}
    \label{eq:traj_prob_expansion}
\end{equation}
where $p(\bm{s}_0)$ represents the distribution of the initial state, and $p(\bm{s}_{t+1}, r_{t+1} | \bm{s}_t, \bm{a}_t)$ denotes the environment's transition dynamics. These terms are independent with the policy’s parameters $\bm{\theta}$ so that their gradients equal to zero. Substituing $G(\tau) = \sum_{t=1}^T r_t$ and Equation~(\ref{eq:traj_prob_expansion}) into Equation~(\ref{eq:pg_gradient_v1}), we obtain
\begin{equation}
    \nabla_{\bm{\theta}} \mathbb{E}_{\tau \sim p_{\bm{\theta}}(\tau)} [G(\tau)] = \mathbb{E}_{\tau \sim p_{\bm{\theta}}(\tau)} \left[ \sum_{t=0}^{T-1} \nabla_{\bm{\theta}} \log \pi(\bm{a}_t|\bm{s}_t; \bm{\theta}) \sum_{t^\prime = 1}^T r_{t^\prime} \right]
\end{equation}

Intuitively, $\log \pi(\bm{a}_t | \bm{s}_t; \bm{\theta})$ describes the log-probability of taking an action $\bm{a}_t$ when observing the state $\bm{s}_t$ under the current policy's parameter $\bm{\theta}$. By multiplying this log-probability with the accumulated rewards and taking the gradient with respect to the $\bm{\theta}$, we are pushing up the log-probabilities of each action in proportion to $G(\tau)$ through a gradient ascent step. However, this approach overlooks the fact that rewards obtained prior to taking an action have no influence on the quality of that action—the rewards that occur after the action are the relevant ones. To address this, people have introduced the concept of causality~\cite{schulman_phd_thesis_2016_berkeley, cs285_berkeley}, which states that the policy cannot influence rewards that have already been obtained. By exploiting this causality, we can modify the objective to account only for future rewards, resulting in the following adjusted form:
\begin{equation}
    \renewcommand{\CancelColor}{\color{blue}} 
    \begin{aligned}
        \nabla_{\bm{\theta}} \mathbb{E}_{\tau \sim p_{\bm{\theta}}(\tau)} [G(\tau)] &= \mathbb{E}_{\tau \sim p_{\bm{\theta}}(\tau)} \left[ \sum_{t=0}^{T-1} \nabla_{\bm{\theta}} \log \pi(\bm{a}_t|\bm{s}_t; \bm{\theta}) \sum_{t^\prime = 1}^T r_{t^\prime} \right] \\
        &= \mathbb{E}_{\tau \sim p_{\bm{\theta}}(\tau)} \left[ \sum_{t=0}^{T-1} \nabla_{\bm{\theta}} \log \pi(\bm{a}_t|\bm{s}_t; \bm{\theta}) \left( \bcancel{\sum_{t^\prime = 1}^t r_{t^\prime}} + \sum_{t^\prime = t+1}^T r_{t^\prime} \right) \right] \\
        &= \mathbb{E}_{\tau \sim p_{\bm{\theta}}(\tau)} \left[ \sum_{t=0}^{T-1} \nabla_{\bm{\theta}} \log \pi(\bm{a}_t|\bm{s}_t; \bm{\theta}) \sum_{t^\prime = t+1}^T r_{t^\prime} \right]
    \end{aligned}
\end{equation}
where the term $\mathbb{E}_{\tau \sim p_{\bm{\theta}}(\tau)} \left[ \sum_{t=0}^{T-1} \nabla_{\bm{\theta}} \log \pi(\bm{a}_t|\bm{s}_t; \bm{\theta}) \sum_{t^\prime = 1}^t r_{t^\prime} \right]$ vanishes due to the causality assumption. Furthermore, it can be shown mathematically that the term $\sum_{t=0}^{T-1} \nabla_{\bm{\theta}} \log \pi(\bm{a}_t|\bm{s}_t; \bm{\theta}) \sum_{t^\prime = 1}^t r_{t^\prime}$ has zero mean but nonzero variance~\cite{openai_spinning_up_2018}. Therefore, leveraging causality is an effective way to reduce variance in policy gradient methods.

It is worth noting that a discount factor $\gamma$ is typically introduced to assign greater importance to immediate rewards. With this, the gradient estimator is given by
\begin{equation}
     \nabla_{\bm{\theta}} \mathbb{E}_{\tau \sim p_{\bm{\theta}}(\tau)} [G(\tau)] = \mathbb{E}_{\tau \sim p_{\bm{\theta}}(\tau)}\left[ \sum_{t=0}^{T-1} \nabla_{\bm{\theta}} \log \pi(\bm{a}_t|\bm{s}_t; \bm{\theta}) \sum_{t^\prime = t+1}^T \gamma^{t^\prime - t} r_{t^\prime} \right]
\end{equation}
In practice, we often collect a batch of trajectories and estimate the gradient of the expected return by averaging over the batch of trajectories. The algorithm is summarized in Algorithm~\ref{alg:mc_policy_gradient}.

\IncMargin{1.2em}

\begin{algorithm}[h]
    \caption{Monte-Carlo Policy Gradient}
    \label{alg:mc_policy_gradient}
    \LinesNumbered


    \ForEach{training iteration}
    {
        1. sampling \
        
        Collect a set of trajectories $\mathcal{D}_k = \{ \tau_i \}$ by playing the current policy $\pi_k$\

        2. policy evaluation \

        Compute the return $G(\tau)$ via Monte-Carlo estimation.
        
        Estimate the policy gradient
        \begin{equation*}
            \nabla J(\bm{\theta}_k) = \frac{1}{|\mathcal{D}_k|} \sum_{\tau \in \mathcal{D}_k} \sum_{t=0}^{T-1} \nabla_{\bm{\theta}_k} \log \pi(\bm{a}_t|\bm{s}_t; \bm{\theta}_k) G(\tau)
        \end{equation*} \

        3. policy improvement \

        $\bm{\theta}_{k+1} \leftarrow \bm{\theta}_{k} + \alpha \nabla J(\bm{\theta}_{k})$ \
    }
\end{algorithm}

\DecMargin{1.2em}

\subsection{Baselines}
We can reduce the variance of the policy gradient estimator by introducing a baseline term $b(\bm{s}_t)$ that have no effect on the expectation.

\begin{equation}
    \nabla_{\bm{\theta}} \mathbb{E}_{\tau \sim p_{\bm{\theta}}(\tau)} [G(\tau)] = \mathbb{E}_{\tau \sim p_{\bm{\theta}}(\tau)} \left[ \sum_{t=0}^{T-1} \nabla_{\bm{\theta}} \log \pi(\bm{a}_t|\bm{s}_t; \bm{\theta}) \left( \sum_{t^\prime = t+1}^T r_{t^\prime} - b(\bm{s_t}) \right) \right]
\end{equation}
A near-optimal choice of baseline is the state-value function~\cite{zhaoshiyu_rlbook}. By the definition of advantage function, we rewrite the objective and its gradient as:
\begin{subequations}
    \begin{align}
        \mathcal{L}^{\text{VPG}}(\bm{\theta}) &= \mathbb{E}_{\tau \sim p_{\bm{\theta}}(\tau)} \left[ A_{\pi_{\bm{\theta}}}(\bm{s}, \bm{a}) \right] \label{eq:loss_vpg} \\
        \nabla_{\bm{\theta}}\mathcal{L}^{\text{VPG}}(\bm{\theta}) &= \mathbb{E}_{\tau \sim p_{\bm{\theta}}(\tau)} \left[ \nabla_{\bm{\theta}} \log \pi(\bm{a} | \bm{s}; \bm{\theta}) A_{\pi_{\bm{\theta}}}(\bm{s}, \bm{a}) \right] \label{eq:grad_of_loss_vpg}
    \end{align}
\end{subequations}

To demonstrate the advantage of introducing a baseline, consider the following example. For a given agent's state $\bm{s}_t$, two different actions, $\bm{a}_t^{[1]}$ and $\bm{a}_t^{[2]}$, lead to distinct trajectories with accumulated rewards $G^{[1]}$ and $G^{[2]}$, respectively. Let $\bar{G}$ denote the average accumulated reward across various actions. Assume $G^{[1]} > \bar{G} > G^{[2]} > 0$. Without a baseline, the gradient ascent step would lead to an increase in the selection probabilities of both actions, with a relatively greater increment for $\bm{a}_t^{[1]}$ compared to  $\bm{a}_t^{[2]}$. However, by introducing a baseline, we calculate the advantages as $A^{[1]} = G^{[1]} - \bar{G} > 0$ and $A^{[2]} = G^{[2]} - \bar{G} < 0$. As a result, the gradient ascent step increases the probability of selecting $\bm{a}_t^{[1]}$ while decreasing the probability of selecting $\bm{a}_t^{[2]}$. This method ensures that policy updates focus on the relative performance of actions compared to the baseline, rather than being influenced by absolute rewards, which can vary widely. When the state-value function is used as the baseline, the policy gradient step increases the probability of actions that are better than average and decreases the probability of those that are worse than average~\cite{gae_iclr2016}. This leads to more stable and focused policy updates, resulting in more efficient learning and improved convergence properties in reinforcement learning algorithms.


\subsection{Actor-Critic Methods}
We know that in policy gradient methods, evaluating a specific policy typically involves two steps: first, estimating the expected return $G(\tau)$ through sampling, where $G(\tau)$ can take various forms, such as $Q_\pi(\bm{s}_t, \bm{a}_t)$ and $A_\pi(\bm{s}_t, \bm{a}_t)$. Then, we compute its gradient with respect to the policy parameters $\bm{\theta}$. This gradient provides information about the direction of the greatest change in the expected return. It is worth noting that in the first step, the expected return can be estimated using both the Monte Carlo method and the Temporal Difference method. If the expected return is estimated using the Monte Carlo method, the resulting algorithm is called the Monte Carlo policy gradient, or REINFORCE, which we have discussed previously. In this section, we will discuss the algorithms where the expected return is estimated using Temporal Difference methods, and the resulting algorithms are often referred to as \textbf{\textit{Actor-Critic methods}}.

Recall that the key idea of TD methods is to leverage previous value estimates when calculating returns, as shown in Equation~(\ref{eq:n_step_return_v}) and Equation~(\ref{eq:n_step_return_q}). In Section~\ref{sec:value_parameterization}, we also introduced the value parameterization technique. Suppose we use a neural network to approximate the state-value function, denoted as $\hat{V}(\bm{s}; \bm{w})$, where $\bm{w}$ represents the network parameters. Using 1-step TD, the expected return is estimated by $r_{t+1} + \gamma \hat{V}(\bm{s}_{t+1}; \bm{w})$. In actor-critic methods, this value network is referred to as the \textbf{\textit{critic}}, while the policy network is known as the \textbf{\textit{actor}}. A basic actor-critic algorithm is presented in Algorithm~\ref{alg:actor_critic}.

It is worth noting that, compared to Monte Carlo policy gradient methods, actor-critic methods have the advantage of not requiring the episode to terminate before updating the policy. Additionally, due to the nature of TD methods, they typically exhibit lower variance.

\IncMargin{1.2em}

\begin{algorithm}[h]
    \caption{Actor-Critic Methods}
    \label{alg:actor_critic}
    \LinesNumbered


    \ForEach{training iteration}
    {
        1. sampling \
        
        Collect a set of trajectories $\mathcal{D}_k = \{ \tau_i \}$ by playing the current policy $\pi_k$\

        2. policy evaluation \

        Compute the return $G(\tau)$ based on current value estimations.
        
        Estimate the policy gradient
        \begin{equation*}
            \nabla J(\bm{\theta}_k) = \frac{1}{|\mathcal{D}_k|} \sum_{\tau \in \mathcal{D}_k} \sum_{t=0}^{T-1} \nabla_{\bm{\theta}_{k}} \log \pi(\bm{a}_t|\bm{s}_t; \bm{\theta}_k) G(\tau)
        \end{equation*} \

        Update the value estimator
        \begin{equation*}
            \bm{w}_{k+1} = \mathop{\arg\min}\limits_{\bm{w}_k} \frac{1}{|\mathcal{D}_k|T} \sum_{\tau \in \mathcal{D}_k} \sum_{t=0}^T \left( r_t + \gamma \hat{V}_\pi(\bm{s}_{t+1}; \bm{w}_k) - \hat{V}_\pi(\bm{s}_t; \bm{w}_k) \right)^2
        \end{equation*}

        3. policy improvement \

        $\bm{\theta}_{k+1} \leftarrow \bm{\theta}_{k} + \alpha \nabla J(\bm{\theta}_{k})$ \
    }
\end{algorithm}

\DecMargin{1.2em}

\subsection{Generalized Advantage Estimation (GAE)}
In this section, we present a widely used technique for estimating the advantage function, which utilizes the $\lambda$-return to compute advantage estimates. Similar to the introduction of TD$(\lambda)$, where TD$(\lambda)$ is formulated as an exponentially weighted average of $n$-step returns, 
\textbf{\textit{Generalized Advantage Estimation (GAE)}}, or GAE($\lambda$)~\cite{gae_iclr2016}, is defined as an exponentially weighted average of $n$-step advantage estimators. 

Consider a trajectory $\tau = \{\bm{s}_t, \bm{a}_t, r_{t+1}\}_{t=0}^{T-1}$. When the agent is in state $\bm{s}_t$ and takes action $\bm{a}_t$, the $n$-step return can be used to estimate the value of the state-action pair $(\bm{s}_t, \bm{a}_t)$, denoted as $\hat{Q}_{t:t+n}$. By definition, the difference between the estimated action value $\hat{Q}_{t:t+n}$ and the state value $\hat{V}_\pi(\bm{s}_t)$ provides a $n$-step advantage estimator, represented as $\hat{A}_{t:t+n}$:
\begin{equation}
    \begin{aligned}
        \hat{A}_{t:t+n} &= \hat{Q}_{t:t+n} - \hat{V}_\pi(\bm{s}_t) \\
        &= r_{t+1} + \gamma r_{t+2} + \cdots + \gamma^{n-1}r_{t+n} + \gamma^n \hat{V}_\pi(\bm{s}_{t+n}) - \hat{V}_\pi(\bm{s}_t)
    \end{aligned}
    \label{eq:n_step_advantage_def}
\end{equation}

The $n$-step advantage estimator can be shown to be equivalent to the discounted sum of TD errors, which can be derived by demonstrating the equivalence between Equation~(\ref{eq:n_step_advantage_as_sum_of_td_errors}) and Equation~(\ref{eq:n_step_advantage_def}). Here, the TD error is defined as $\delta_{t+1}^V = r_{t+1} + \gamma \hat{V}_\pi(\bm{s}_{t+1}) - \hat{V}_\pi(\bm{s}_t)$:
\begin{subequations}
    \begin{align}
        \hat{A}_{t:t+n} &= \sum_{l=1}^{n} \gamma^{l-1} \delta_{t+l}^V \label{eq:n_step_advantage_as_sum_of_td_errors}\\
        &= \delta_{t+1}^V + \gamma \delta_{t+2}^V + \cdots + \gamma^{n-1} \delta_{t+n}^V \notag \\
        &= r_{t+1} + \gamma {\color{blue}\hat{V}_\pi (\bm{s}_{t+1})} - \hat{V}_\pi(\bm{s}_t) + \gamma \left( r_{t+2} + \gamma {\color{blue}\hat{V}_\pi(\bm{s}_{t+2})} - {\color{magenta}\hat{V}_\pi(\bm{s}_{t+1})} \right) \notag\\
        &\quad \quad + \gamma^2 \left( r_{t+3} + \gamma {\color{blue}\hat{V}_\pi(\bm{s}_{t+3})} - {\color{magenta}\hat{V}_\pi(\bm{s}_{t+2})} \right) + \cdots + \gamma^{n-1} \left( r_{t+n} + \gamma \hat{V}_\pi(\bm{s}_{t+n}) - {\color{magenta}\hat{V}_\pi(\bm{s}_{t+n-1})} \right) \notag\\
        &= r_{t+1} + \gamma {\color{blue}\hat{V}_\pi(\bm{s}_{t+1})} - \gamma {\color{magenta}\hat{V}_\pi(\bm{s}_{t+1})} + \gamma r_{t+2} + \gamma^2 {\color{blue}\hat{V}_\pi(\bm{s}_{t+2})} - \gamma^2 {\color{magenta}\hat{V}_\pi(\bm{s}_{t+2})} \notag\\
        &\quad \quad + \cdots + \gamma^{n-1} r_{t+n} + \gamma^n \hat{V}_\pi(\bm{s}_{t+n}) - \hat{V}_\pi(\bm{s}_t) \notag\\
        &= r_{t+1} + \gamma r_{t+2} + \cdots + \gamma^{n-1}r_{t+n} + \gamma^n \hat{V}_\pi(\bm{s}_{t+n}) - \hat{V}_\pi(\bm{s}_t) \notag
    \end{align}
\end{subequations}

GAE($\lambda$) is defined as the exponentially-weighted average of the $n$-step advantage estimators:
\begin{equation}
    \begin{aligned}
        A_t^\lambda &= (1 - \lambda)\left( A_{t:t+1} + \lambda A_{t:t+2} + \lambda^2 A_{t:t+3} + \cdots \right) \\
        &= (1 - \lambda)\left( \delta_{t+1}^V + \lambda \left( \delta_{t+1}^V + \gamma \delta_{t+2}^V \right) + \lambda^2\left( \delta_{t+1}^V + \gamma \delta_{t+2}^V + \gamma^2 \delta_{t+3}^V \right) + \cdots\right) \\
        &= (1 - \lambda)\left[ \delta_{t+1}^V (1 + \lambda + \lambda^2 + \cdots) + \gamma \lambda \delta_{t+2}^V (1 + \lambda + \lambda^2 + \cdots) + \gamma^2 \lambda^2 \delta_{t+3}^V (1 + \lambda + \lambda^2 + \cdots) + \cdots \right] \\
        &= (1 - \lambda)\left[ \delta_{t+1}^V \left(\frac{1}{1-\lambda}\right) + \gamma \lambda \delta_{t+2}^V \left(\frac{1}{1-\lambda}\right) + \gamma^2 \lambda^2 \delta_{t+3}^V \left(\frac{1}{1-\lambda}\right) + \cdots\right] \\
        &= \sum_{l=1}^{\infty} (\gamma \lambda)^{l-1} \delta_{t+l}^V
    \end{aligned}
\end{equation}
where we observe two special cases by setting $\lambda = 0$ and $\lambda = 1$:
\begin{subequations}
    \begin{align}
        \lambda &= 0: \quad A_t^0 = \delta_{t+1}^V = r_{t+1} + \gamma \hat{V}_\pi(\bm{s}_{t+1}) - \hat{V}_\pi(\bm{s}_t) \\
        \lambda &= 1: \quad A_t^1 = \sum_{l=1}^\infty \gamma^{l-1} \delta_{t+l}^V = \sum_{l=1}^\infty \gamma^{l-1} r_{t+l} - \hat{V}_\pi(\bm{s}_t)
    \end{align}
\end{subequations}
When $\lambda = 1$, GAE($\lambda$) corresponds to an $\infty$-step advantage estimator, equivalent to the Monte Carlo method, which is characterized by high variance due to the summation of multiple terms. Conversely, when $\lambda = 0$, GAE($\lambda$) reduces to the 1-step TD residual, which has lower variance but introduces higher bias. For intermediate values $0 < \lambda < 1$, GAE($\lambda$) strikes a balance between bias and variance, with the trade-off controlled by the parameter $\lambda$.







\subsection{Policy Gradient with Off-Policy Samples}

\begin{figure}[htbp]
    \centering
    \includegraphics[width=\textwidth, trim =5 275 10 110, clip]{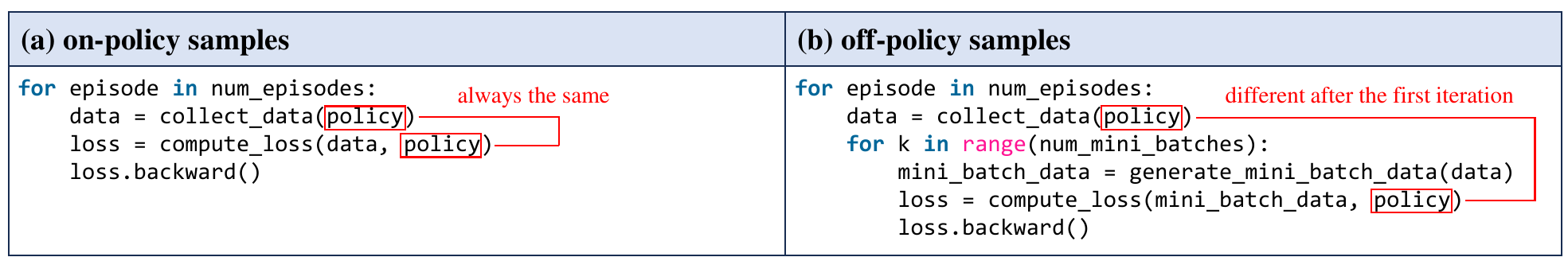}

    \caption{\textbf{Illustration of the differences in training using on-policy and off-policy samples.} (a) When the policy used for data collection is the same as the policy used for loss computation, the training is referred to as using on-policy samples. (b) When data is collected using policy $\pi$ and $k$ policy update steps are performed with mini-batch data, the policy used for loss computation after the first update step differs from the policy used for data collection. This training process is referred to as using off-policy samples.}

    \label{fig:off_policy_sample_illustration}
\end{figure}

In previous sections, we estimated the gradient of the expected return with respect to the current policy parameters, as illustrated in Figure~\ref{fig:off_policy_sample_illustration}(a). This approach is referred to as using \textbf{\textit{on-policy samples}}, where the policy used for sampling and evaluation is the same. However, this training procedure has a significant limitation: samples collected using the current policy can only be used for a single gradient step. Afterward, a new batch of samples must be collected to perform the gradient step for the updated policy. This process results in low data efficiency. To address this issue, we can leverage \textbf{\textit{off-policy samples}}, as illustrated in Figure~\ref{fig:off_policy_sample_illustration}(b). Instead of using the entire batch of collected data for just one gradient step, off-policy training allows us to reuse the same batch by generating mini-batches for multiple gradient updates. Let the policy used for sampling be denoted as $\pi(\bm{a} | \bm{s}; {\color{blue} \bm{\theta}_{\text{old}}})$ and the policy used to perform the $k$-th gradient step as $\pi(\bm{a} | \bm{s}; {\color{red} \bm{\theta}_k})$. At each gradient step, we compute: 
\begin{equation}
    \nabla_{{\color{red} \bm{\theta}_k}} \mathbb{E}_{\tau \sim p_{{\color{blue} \bm{\theta}_{\text{old}}}}(\tau)} = \mathbb{E}_{\tau \sim p_{{\color{blue} \bm{\theta}_{\text{old}}}}(\tau)} \left[ \sum_{t=0}^{T-1} \nabla_{\bm{\theta}} \log \pi(\bm{a}_t|\bm{s}_t; {\color{red} \bm{\theta}_k}) A_{\pi_{{\color{red} \bm{\theta}_k}}}(\bm{s}_t, \bm{a}_t) \right],
\end{equation}
where ${\color{red} \bm{\theta}_k} \neq {\color{blue} \bm{\theta}_{\text{old}}}$ after the first gradient step for each episode. The challenge arises because the expectation must now be computed over the sampling distribution $p_{{\color{blue} \bm{\theta}_{\text{old}}}}(\tau)$, even though the policy being updated is $\pi(\bm{a}|\bm{s}; {\color{red} \bm{\theta}_k})$.
To resolve this, the remaining part of this section introduces a technique called \textbf{\textit{importance sampling}}, which enables us to compute expectations when the sampling distribution differs from the distribution under policy evaluation.

Importance sampling is a general technique for estimating the expectation of a function $f(\bm{x})$ of a random variable $\bm{x}$ under one probability distribution $p(\bm{x})$, using samples drawn from a different distribution $q(\bm{x})$. We begin our detailed introduction by considering the problem of estimating $\mathbb{E}_{\bm{x} \sim p(\bm{x})}[f(\bm{x})]$, the expectation of $f(\bm{x})$ with respect to $\bm{x} \sim p(\bm{x})$, where $p(\bm{x})$ is the true distribution of $\bm{x}$. A straightforward approach would be to generate independent and identically distributed (i.i.d.) samples from $p(\bm{x})$ and compute their average to obtain an unbiased estimate of $\mathbb{E}_{\bm{x} \sim p(\bm{x})}[f(\bm{x})]$. However, this approach is not feasible when $p(\bm{x})$ is unknown and we only have i.i.d. samples from another distribution $q(\bm{x})$. In this case, importance sampling can be applied as follows:
\begin{equation}
    \begin{aligned}
        \mathbb{E}_{\bm{x} \sim p(\bm{x})}[f(\bm{x})] &= \int p(\bm{x}) f(\bm{x}) \text{d}\bm{x} \\
        &= \int p(\bm{x}) \frac{q(\bm{x})}{q(\bm{x})} f(\bm{x}) \text{d}\bm{x} \\
        &= \int q(\bm{x}) \left[ \frac{p(\bm{x})}{q(\bm{x})} f(\bm{x}) \right] \text{d}\bm{x} \\
        &= \mathbb{E}_{\bm{x} \sim q(\bm{x})}\left[ \frac{p(\bm{x})}{q(\bm{x})} \bm{x} \right]
    \end{aligned}
\end{equation}


Using the importance sampling technique, we can conveniently compute the objective and its gradient when leveraging off-policy samples:
\begin{subequations}
    \begin{align}
        \mathcal{L}^{\text{IS}}(\bm{\theta}) &= \mathbb{E}_{\tau \sim p_{\bm{\theta}_\text{old}}(\tau)} \left[ \frac{\pi(\bm{a}|\bm{s};\bm{\theta})}{\pi(\bm{a}|\bm{s};\bm{\theta}_\text{old})} A_{\pi_{\bm{\theta}_{\text{old}}}}(\bm{s}, \bm{a}) \right] \label{eq:loss_is}\\
        \nabla_{\bm{\theta}} \mathcal{L}^{\text{IS}}(\bm{\theta}) &= \mathbb{E}_{\tau \sim p_{\bm{\theta}_\text{old}}(\tau)} \left[ \frac{\pi(\bm{a}|\bm{s};\bm{\theta})}{\pi(\bm{a}|\bm{s};\bm{\theta}_\text{old})} \nabla_{\bm{\theta}} \log \pi(\bm{a} | \bm{s}; \bm{\theta}) A_{\pi_{\bm{\theta}_{\text{old}}}}(\bm{s}, \bm{a}) \right], \label{eq:gradient_of_loss_is}
    \end{align}
\end{subequations}
where $\mathcal{L}^{\text{IS}}(\bm{\theta})$ represents the loss function for policy gradient incorporating the importance sampling technique. Equations~(\ref{eq:loss_is}) and (\ref{eq:gradient_of_loss_is}) are derived by directly applying the importance sampling technique to Equations~(\ref{eq:loss_vpg}) and (\ref{eq:grad_of_loss_vpg}), respectively. Furthermore, Equation~(\ref{eq:gradient_of_loss_is}) can be derived by directly taking the gradient of Equation~(\ref{eq:loss_is}):
\begin{equation}
    \begin{aligned}
        \nabla_{\bm{\theta}} \mathcal{L}^{\text{IS}}(\bm{\theta}) &= \nabla_{\bm{\theta}} \mathbb{E}_{\tau \sim p_{\bm{\theta}_\text{old}}(\tau)} \left[ \frac{\pi(\bm{a}|\bm{s};\bm{\theta})}{\pi(\bm{a}|\bm{s};\bm{\theta}_\text{old})} A_{\pi_{\bm{\theta}_{\text{old}}}}(\bm{s}, \bm{a}) \right] \\
        &= \mathbb{E}_{\tau \sim p_{\bm{\theta}_\text{old}}(\tau)} \left[ \frac{\nabla_{\bm{\theta}}\pi(\bm{a}|\bm{s};\bm{\theta})}{\pi(\bm{a}|\bm{s};\bm{\theta}_\text{old})} A_{\pi_{\bm{\theta}_{\text{old}}}}(\bm{s}, \bm{a}) \right] \\
        &= \mathbb{E}_{\tau \sim p_{\bm{\theta}_\text{old}}(\tau)} \left[ \frac{\pi(\bm{a} | \bm{s}; \bm{\theta})}{\pi(\bm{a} | \bm{s}; \bm{\theta})} \frac{\nabla_{\bm{\theta}}\pi(\bm{a}|\bm{s};\bm{\theta})}{\pi(\bm{a}|\bm{s};\bm{\theta}_\text{old})} A_{\pi_{\bm{\theta}_{\text{old}}}}(\bm{s}, \bm{a}) \right] \\
        &= \mathbb{E}_{\tau \sim p_{\bm{\theta}_\text{old}}(\tau)} \left[ \frac{\pi(\bm{a} | \bm{s}; \bm{\theta})}{\pi(\bm{a} | \bm{s}; \bm{\theta}_\text{old})} \frac{\nabla_{\bm{\theta}}\pi(\bm{a}|\bm{s};\bm{\theta})}{\pi(\bm{a}|\bm{s};\bm{\theta})} A_{\pi_{\bm{\theta}_{\text{old}}}}(\bm{s}, \bm{a}) \right] \\
        &= \mathbb{E}_{\tau \sim p_{\bm{\theta}_\text{old}}(\tau)} \left[ \frac{\pi(\bm{a} | \bm{s}; \bm{\theta})}{\pi(\bm{a} | \bm{s}; \bm{\theta}_\text{old})} \nabla_{\bm{\theta}}\log\pi(\bm{a} | \bm{s}; \bm{\theta}) A_{\pi_{\bm{\theta}_{\text{old}}}}(\bm{s}, \bm{a}) \right].
    \end{aligned}
\end{equation}

By leveraging the importance sampling technique, the same batch of data can be reused for multiple gradient steps, thereby significantly improving data efficiency.

\subsection{Proximal Policy Optimization (PPO)}

Empirically, the policy gradient methods discussed so far often suffer from instability, where even a single poorly chosen update step can significantly degrade the performance. The \textbf{\textit{Proximal Policy Optimization (PPO)}} algorithm addresses this limitation by introducing a clipped objective function that restricts the magnitude of policy updates. This mechanism prevents excessive changes to the policy, ensuring updates remain controlled and the policy evolves incrementally. By stabilizing the training process, PPO reduces the risk of divergence and enhances overall robustness. It has become one of the most popular deep reinforcement algorithms in diverse applications. 

PPO optimizes a clipped loss defined as:
\begin{equation}
    \mathcal{L}^\text{CLIP} = \min \left( \frac{\pi(\bm{a}|\bm{s};\bm{\theta})}{\pi(\bm{a}|\bm{s};\bm{\theta}_\text{old})} A_{\pi_{\bm{\theta}_{\text{old}}}}(\bm{s}, \bm{a}), {\color{blue} \text{clip}\left( \frac{\pi(\bm{a}|\bm{s};\bm{\theta})}{\pi(\bm{a}|\bm{s};\bm{\theta}_\text{old})}, 1 - \epsilon, 1 + \epsilon \right) A_{\pi_{\bm{\theta}_{\text{old}}}}(\bm{s}, \bm{a})} \right)
    \label{eq:loss_clip}
\end{equation}
where the $\text{clip}(\cdot)$ operator constrains the ratio $\frac{\pi(\bm{a}|\bm{s};\bm{\theta})}{\pi(\bm{a}|\bm{s};\bm{\theta}_\text{old})}$ to lie within the range $[1 - \epsilon, 1 + \epsilon]$, and $\epsilon$ is a hyperparameter that controls the extent of the policy update. Although the above equation appears complex, its intuition becomes clearer through its simplified version:
\begin{equation}
    \mathcal{L}^\text{CLIP} = \left\{ \begin{aligned}
        &\mathbb{E}_{\tau \sim p_{\bm{\theta}_\text{old}}(\tau)} \left[ \min \left( \frac{\pi(\bm{a}|\bm{s};\bm{\theta})}{\pi(\bm{a}|\bm{s};\bm{\theta}_\text{old})}, 1 + \epsilon \right) A_{\pi_{\bm{\theta}_{\text{old}}}}(\bm{s}, \bm{a}) \right] & \text{if }A_{\pi_{\bm{\theta}_{\text{old}}}}(\bm{s}, \bm{a}) \geq 0 \\
        &\mathbb{E}_{\tau \sim p_{\bm{\theta}_\text{old}}(\tau)} \left[ \max \left( \frac{\pi(\bm{a}|\bm{s};\bm{\theta})}{\pi(\bm{a}|\bm{s};\bm{\theta}_\text{old})}, 1 - \epsilon \right) A_{\pi_{\bm{\theta}_{\text{old}}}}(\bm{s}, \bm{a}) \right] & \text{if }A_{\pi_{\bm{\theta}_{\text{old}}}}(\bm{s}, \bm{a}) < 0
    \end{aligned} \right.
\end{equation}
We offer the following intuitive explanation based on the simplified form of the objective:
\begin{itemize}
    \item when $A_{\pi_{\bm{\theta}_{\text{old}}}}(\bm{s}, \bm{a}) \geq 0$, 
    it implies that the specific action $\bm{a}$ is at least as good as the average action under the state $\bm{s}$. In this case, we aim to make that action more likely to be chosen, i.e., increase $\pi(\bm{a}|\bm{s}; \bm{\theta})$, to improve the objective. However, the use of the min operator imposes a limit on how much the objective can increase. Once $\pi(\bm{a}|\bm{s};\bm{\theta})> (1 + \epsilon)\pi(\bm{a}|\bm{s};\bm{\theta}_\text{old})$, the min term activates, and this part of the objective reaches a maximum of $(1 + \epsilon)\pi(\bm{a}|\bm{s};\bm{\theta}_\text{old})$.
    \item when $A_{\pi_{\bm{\theta}_{\text{old}}}}(\bm{s}, \bm{a}) < 0$, it indicates that the specific action $\bm{a}$ is worse than the average action under the state $\bm{s}$. In this scenario, we aim to make that action less likely to be chosen, i.e., reduce $\pi(\bm{a}|\bm{s}; \bm{\theta})$, to improve the objective. However, the max operator imposes a limit on how much the objective can increase. Once $\pi(\bm{a}|\bm{s};\bm{\theta})> (1 + \epsilon)\pi(\bm{a}|\bm{s};\bm{\theta}_\text{old})$,  the max term activates, and this part of the objective reaches a floor of $(1 + \epsilon)\pi(\bm{a}|\bm{s};\bm{\theta}_\text{old})$. 
\end{itemize}

In practice, we employ an actor-critic approach, utilizing a learned state-value function to compute variance-reduced advantage estimates. To encourage sufficient exploration, the overall objective is further augmented with an entropy bonus term. The detailed procedure is presented in Algorithm~\ref{alg:ppo}.

\IncMargin{1.2em}

\begin{algorithm}[h]
    \caption{Proximal Policy Optimization}
    \label{alg:ppo}
    \LinesNumbered
    


    \ForEach{training iteration}
    {
        1. sampling \
        
        Collect a set of trajectories $\mathcal{D}_k = \{ \tau_i \}$ by playing the current policy $\pi_k$\

        2. policy evaluation \

        Compute the return $G(\tau)$ based on current value estimations. \

        Compute advantage estimates using GAE, based on the current value function $\hat{V}_\pi(\bm{s}; \bm{w}_k)$. \
        
        Estimate the policy gradient using the clipped objective:
        \begin{equation*}
            \nabla J(\bm{\theta}_k) = \frac{1}{|\mathcal{D}_k|} \sum_{\tau \in \mathcal{D}_k} \sum_{t=0}^{T-1} \nabla_{\bm{\theta}_{k}} \min \left( \frac{\pi(\bm{a}|\bm{s};\bm{\theta})}{\pi(\bm{a}|\bm{s};\bm{\theta}_k)} A_{\pi_{\bm{\theta}_k}}(\bm{s}, \bm{a}), g(\epsilon, A_{\pi_{\bm{\theta}_k}}(\bm{s}, \bm{a})) \right)
        \end{equation*} \

        Update the value estimator by regression on mean-square error:
        \begin{equation*}
            \bm{w}_{k+1} = \mathop{\arg\min}\limits_{\bm{w}_k} \frac{1}{|\mathcal{D}_k|T} \sum_{\tau \in \mathcal{D}_k} \sum_{t=0}^T \left( r_t + \gamma \hat{V}_\pi(\bm{s}_{t+1}; \bm{w}_k) - \hat{V}_\pi(\bm{s}_t; \bm{w}_k) \right)^2
        \end{equation*}

        3. policy improvement \

        $\bm{\theta}_{k+1} \leftarrow \bm{\theta}_{k} + \alpha \nabla J(\bm{\theta}_{k})$ \
    }
\end{algorithm}

\DecMargin{1.2em}
\section{Conclusion}
In this tutorial, we presented a concise and practical introduction to deep reinforcement learning (DRL), focusing on the Proximal Policy Optimization (PPO) algorithm as a representative and widely adopted method in the field. To help readers efficiently navigate the complexities of DRL, we organized all algorithms under the Generalized Policy Iteration (GPI) framework, providing a unified and systematic understanding of various approaches. By emphasizing intuitive explanations, illustrative examples, and practical implementation techniques over lengthy theoretical proofs, this work offers an accessible and application-oriented learning path for beginners and practitioners.

Through the discussions of value estimation methods, policy gradient techniques, and Generalized Advantage Estimation (GAE), we highlighted the essential components of the PPO algorithm and demonstrated how they interconnect within the GPI framework. This tutorial not only clarifies the conceptual foundations of DRL but also equips readers with practical knowledge and tools for applying these algorithms to real-world problems.


	

\clearpage


\bibliography{ref}

\appendix

\newpage

\section{Implementation Details of Policy Gradient Methods}

\subsection*{A.1~~ Value function clipping}
Standard implementation fits the critic with a clipped objective:
\begin{equation*}
    \mathcal{L}^V = \max \left[ \left(\hat{V}_\pi(\bm{s})-{V}_\text{targ}\right)^2, \left(\text{clip}\left( \hat{V}_\pi(\bm{s}), \hat{V}_\pi(\bm{s}) - \epsilon, \hat{V}_\pi(\bm{s}) + \epsilon \right)-{V}_\text{targ}\right)^2 \right]
    \label{eq:value_loss_clip}
\end{equation*}
where $\hat{V}_\pi(\bm{s})$ is clipped around the previous value estimates and $\epsilon$ is fixed to the same value as the value used to clip probability ratios in \ref{eq:loss_clip}.

\subsection*{A.2~~ Timeout bootstrap}
For continuous tasks, such as robotic control, there is a clear failure termination criterion but no explicit success criterion. During training, a maximum episode length is often imposed, causing the interaction between the agent and the environment to terminate when this limit is reached. This termination is referred to as a timeout. To ensure the value estimation aligns with the original task objectives, the rewards associated with timeout states should be appropriately handled:
\begin{equation*}
    r_T = \left\{ \begin{aligned}
        &r_T & \text{coincides with failure} \\
        &r_T + \gamma\hat{V}_\pi(\bm{s}_{T+1}) & \text{simply timeout}
    \end{aligned} \right.
\end{equation*}
In certain cases where $\bm{s}_{T+1}$ cannot be obtained due to environment resets, $\bm{s}_{T}$ can be used as an approximation.

\subsection*{A.3~~ Advantage Normalization}
To reduce the variance of policy gradients and improve training stability, batch normalization can be applied to the advantage function:
\begin{equation*}
    \hat{A}_\pi \leftarrow \frac{\hat{A}_\pi - \text{mean}\left(\hat{A}_\pi\right)}{\text{std}\left(\hat{A}_\pi\right) + \epsilon}
\end{equation*}

\subsection*{A.4~~ Observation Normalization}
Applying batch normalization to neural network inputs can improve training performance in supervised learning. However, in reinforcement learning, batch normalization is generally not suitable. Instead, a dynamic mean and variance of the states are typically maintained to normalize inputs during training:
\begin{equation*}
\begin{aligned}
    \mu &\gets \mu + \alpha \cdot (\hat{\mu} - \mu) \\
    \sigma^2 &\gets \sigma^2 + \alpha \cdot (\hat{\sigma}^2 - \sigma^2 + \hat{\mu} - \mu \cdot \hat{\mu}) \\
    s &\gets \frac{s - \mu}{\sqrt{\sigma^2} + \epsilon}
\end{aligned}
\end{equation*}
$\hat{\mu}$ and $\hat{\sigma}^2$ represent the mean and variance of the current input batch.

\subsection*{A.5~~ Policy Entropy}
To maintain policy stochasticity for continued exploration, an entropy bonus term can be incorporated into the loss function. Thus, the complete loss function is defined as:
\begin{equation*}
    \mathcal{L} = \mathcal{L}^\text{CLIP}+c_{1}\mathcal{L}^V-c_{2}\mathrm{H}(\pi(\bm{s}))
    \label{eq:complete_ppo_loss}
\end{equation*}
which maximizes an entropy bonus term.

\subsection*{A.6~~ Adaptive Learning Rate}
To achieve more efficient training, the learning rate can be dynamically adjusted based on changes in the KL divergence throughout the training process. It implements a dynamic strategy for controlling the learning rate $\alpha$ through a comparative analysis of the current KL divergence $kl$ against a target threshold $kl^*$.


\subsection*{A.7~~ Gradient clip}
Gradient clipping is a technique introduced to prevent gradient explosion during training. This trick serves to stabilize the training process:
\begin{equation*}
\text{ if } ||\textbf{g}||  > v \text{ then } \textbf{g} \leftarrow {v}\frac{\textbf{g}}{||\textbf{g}||}
\end{equation*}

\end{document}